\documentclass[journal]{IEEEtai}

\usepackage[colorlinks,urlcolor=blue,linkcolor=blue,citecolor=blue]{hyperref}

\usepackage{color,array}

\usepackage{graphicx}

\usepackage{algorithm}
\usepackage{algpseudocode}
\usepackage{diagbox}
\usepackage{subfigure}
\usepackage{amsmath}
\usepackage{amsfonts} 

\usepackage{color,soul}
\usepackage[table]{xcolor}

\begin{document}

\title{Multistream Gaze Estimation with Anatomical Eye Region Isolation by Synthetic to Real\\Transfer Learning}

\author{Zunayed Mahmud, Paul Hungler, Ali Etemad,~\IEEEmembership{Senior Member,~IEEE}
\IEEEcompsocitemizethanks{\IEEEcompsocthanksitem Z. Mahmud, and A. Etemad are with the Department of Electrical and Computer Engineering and Ingenuity Labs Research Institute, Queen's University, Kingston, Ontario, Canada.
	E-mails: \{zunayed.mahmud, ali.etemad\}@queensu.ca
	\IEEEcompsocthanksitem P. Hungler is with Ingenuity Labs Research Institute, Queen's University, Kingston, Ontario, Canada.
	E-mails: {paul.hungler}@queensu.ca}% <-this % stops an unwanted space
% \thanks{Manuscript received April 19, 2005; revised August 26, 2015.}
}

\markboth{}%IEEE Transactions on Artificial Intelligence
{}

\maketitle

\begin{abstract}
We propose a novel neural pipeline, MSGazeNet, that learns gaze representations by taking advantage of the eye anatomy information through a multistream framework. Our proposed solution comprises two components, first a network for isolating anatomical eye regions, and a second network for multistream gaze estimation. The eye region isolation is performed with a U-Net style network which we train using a synthetic dataset that contains eye region masks for the visible eyeball and the iris region. The synthetic dataset used in this stage is \textcolor{black}{procured using the UnityEyes simulator, and consists of 80,000 eye images.} Successive to training, the eye region isolation network is then transferred to the real domain for generating masks for the real-world eye images. In order to successfully make the transfer, we exploit domain randomization in the training process, which allows for the synthetic images to benefit from a larger variance with the help of augmentations that resemble artifacts. 
The generated eye region masks along with the raw eye images are then used together as a multistream input to our gaze estimation network, \textcolor{black}{which consists of wide residual blocks. The output embeddings from these encoders are fused in the channel dimension before feeding into the gaze regression layers.} \textcolor{black}{We evaluate our framework on three gaze estimation datasets and achieve strong performances. Our method surpasses the state-of-the-art by 7.57\% and 1.85\% on two datasets, and obtains competitive results on the other.}
We also study the robustness of our method with respect to the noise in the data and demonstrate that our model is less sensitive to noisy data. Lastly, we perform a variety of experiments including ablation studies to evaluate the contribution of different components and design choices in our solution.
\end{abstract}

\begin{IEEEImpStatement}
Gaze patterns can reveal meaningful information about a person's behaviour and mental state and is often utilized by modern intelligent interactive systems to better understand the users. Gaze can also be a useful communication cue for people with disabilities. The application of gaze estimation ranges from studying human behaviour and psychology to analyzing visual attention in autonomous driving, virtual reality, and remote classrooms. Many of these applications are sensitive to precision and lack user-specific calibration data. In this work, we aim to improve person-independent gaze estimation by presenting a novel framework that integrates eye region segmentation with multistream gaze estimation. Our experiments reveal that using anatomical features in the form of binary masks improves the accuracy of gaze estimation. Our model does not require any calibration samples yet can estimate gaze for unseen users with high accuracy \textcolor{black}{and can be seamlessly integrated into real-time systems.} \textcolor{black}{Since gaze tracking involves the use of an individual's eye image and has the potential to disclose sensitive details about where the user is looking, it is important to first obtain consent and ensure the maintenance of privacy before proceeding with the work.
}
\end{IEEEImpStatement}


\section{Introduction}

\IEEEPARstart{E}{ye} gaze patterns
can be used to characterize important eye-related movement events such as fixation, saccade, and smooth pursuit \cite{smooth}, which in turn can reveal
meaningful information about human behaviour such as a person's emotion, intention, desire, and state of mind. 
This classified eye movement data can further be exploited as useful features in cognitive load detection \cite{cogload}, mental fatigue detection \cite{fatigue}, and stress level analysis \cite{stress}. The application of gaze extends to many different fields such as human-computer interaction (HCI) \cite{hci}, human-robot interaction (HRI) \cite{hri}, visual attention analysis \cite{visatt}, augmented or virtual reality (AR/VR) systems \cite{ar, vr},  autonomous driving \cite{auto}, and others.  
Hence, gaze estimation has become a widely acknowledged research area in computer vision due to its relevance and numerous contributions to various applications. 

Early works on image-based gaze estimation were performed under constrained settings such as fixed head pose and unchanged illumination \cite{uncons1,uncons2}. Subsequently, with newer datasets such as UTMultiview \cite{utm}, Eyediap \cite{eyediap}, and MPIIGaze \cite{mpii}, some of the above-mentioned limitations were mitigated to make gaze estimation more realistic and compatible with in-the-wild scenarios. Such datasets were collected either in a laboratory environment \cite{utm, eyediap} or in daily life settings \cite{mpii}, offering continuous head pose, continuous gaze targets, variation in appearance, and illumination. These datasets are inherently more challenging for gaze estimation given the dynamic nature and variations in experimental conditions. To overcome these challenges, most recent methods \cite{mpii,apple,kaist,clgm} have leveraged deep convolutional neural networks (CNNs) as they are comparatively more robust towards noise and changes in visual factors.

Deep learning solutions for gaze estimation \cite{lenet, mpii, clgm} generally focus on regressing gaze angles directly from the raw eye image, and often do not consider additional information which may be found in different regions of the eye. For instance, the iris region contains important information that can aid in better estimation of gaze, if it were explicitly learned by the network. Moreover, distinct anatomical regions of the eye, for example the visible eyeball and the iris, are not highly complex to detect, making them insensitive to noise and illumination variations throughout the image, and thus highly beneficial for gaze learning. Yet, to the best of our knowledge, gaze estimation solutions have rarely focused on these properties, and explicit detection and learning of anatomical eye regions has so far been ignored. Eye landmarks have in fact been used to help gaze estimation \cite{ETRA, flame}, but are difficult to learn, especially in noisy settings, and considerably increase the dimensionality of the data.

\begin{figure}[t]
\centerline{\includegraphics[width=1\columnwidth]{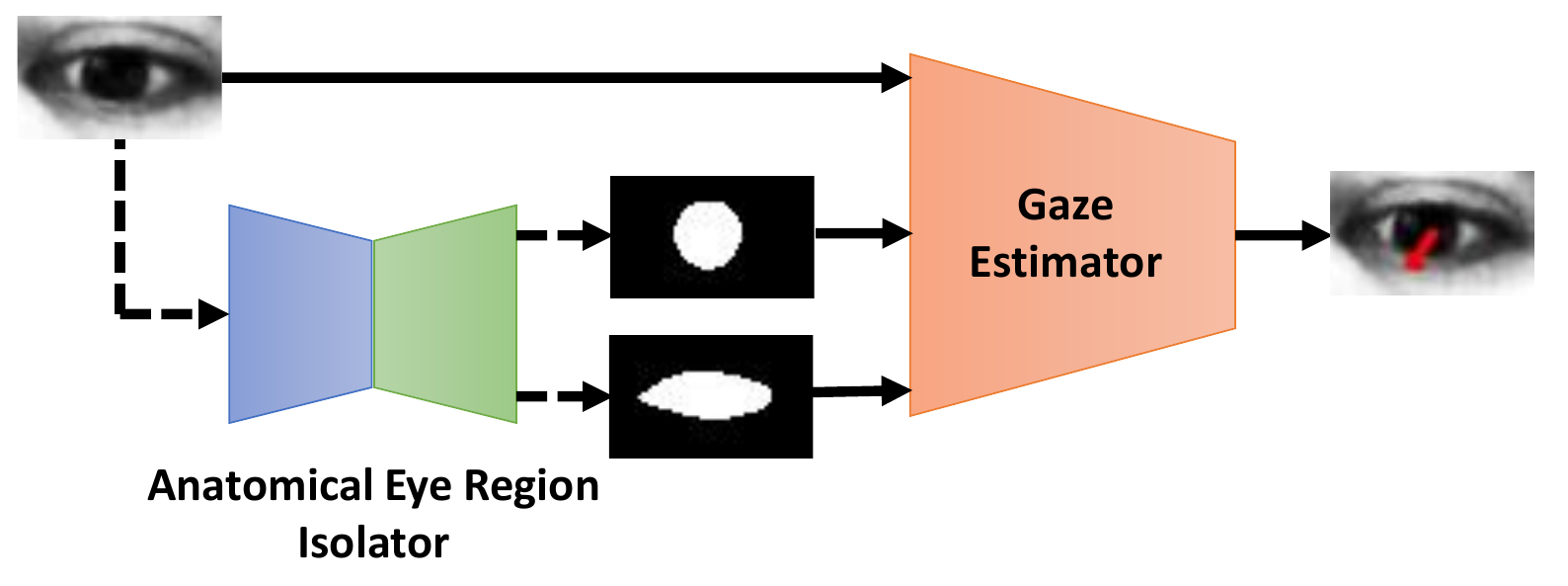}}
\caption{Overview of our proposed method. The anatomical eye region isolation module is used to generate binary eye region masks for the real-world eye images. These masks along with the raw eye image are then used by the multistream gaze estimator to perform gaze estimation. The pre-trained mask generation is represented via dotted lines and the solid lines represent the training pipeline.}
\label{fig:intro}
\end{figure}

In this paper, we aim to improve person-independent gaze estimation by exploiting additional information extracted from raw input images. 
Our proposed solution consists of two key steps, namely anatomical eye region isolation and multistream gaze estimation, together called MSGazeNet. An overview of our proposed framework is depicted in Figure \ref{fig:intro}. Our model first uses a U-Net style network \cite{unet} to perform anatomical eye region isolation, outputting binary masks for two key eye regions which are the iris and the visible eyeball. Following, our model uses a multistream gaze estimation network that takes the raw input eye images along with the outputs of the U-Net, as inputs to estimate the gaze. This component of MSGazeNet uses an encoder in each stream to learn effective gaze-related representations. Next, channel-wise feature fusion is performed. Finally, the fused representations are passed through additional convolutional layers and a regression block consisting a set of fully connected (FC) layers to estimate the output gaze. \textcolor{black}{Given that the existing real-world datasets that are used for gaze estimation do not contain detailed eye structure information such as visible eye region or the iris mask, we train the anatomical eye region isolation network exclusively on a synthetic dataset which we \textcolor{black}{procure} using UnityEyes simulator \cite{unityeyes}.} \textcolor{black}{This synthetic dataset consists of 80,000} synthetic eye images and corresponding eye region masks. Once the isolation network is trained \textcolor{black}{in the synthetic domain}, we perform transfer learning and integrate it into MSGazeNet. We \textcolor{black}{also} perform domain randomization when training the isolation network to ensure that the domain gap between the synthetic images and the real images used to train the downstream network is reduced. The weights of the \textcolor{black}{isolation} network are then frozen and the rest of the network is trained on the real-world gaze datasets for gaze estimation. We use three publicly available gaze datasets, MPIIGaze \cite{mpii}, Eyediap \cite{eyediap}, and UTMultiview \cite{utm}, to evaluate our solution. MSGazeNet obtains strong results, outperforming the state-of-the-art on Eyediap and UTMultiview, and achieving competitive results on MPIIGaze. We also perform a number of ablation studies and qualitative analysis to test the impact of different components and parameters in our network. Lastly, we perform a robustness analysis to investigate the performance across different amount of noise existing in the real-world datasets, and observe that MSGazeNet performs more robustly in comparison to prior works in this area.

Our contributions in this paper are \textcolor{black}{three}-fold.
    \textcolor{black}{(\textbf{1}) We propose a novel deep neural framework for gaze estimation which consists of two main components, an anatomical eye region isolation module, followed by a gaze estimation network, together called MSGazeNet. The isolation network detects key areas of the eye which are informative toward gaze estimation. Our approach eliminates the need for additional inputs such as head pose and eye landmarks, which are often required by multimodal solutions in the area.}
    \textcolor{black}{(\textbf{2}) The eye region isolation network is solely trained using a synthetic dataset that contains over 80,000 eye images along with their iris and visible eyeball masks. The trained network is used to extract binary masks for real-world eye images. We perform domain randomization utilizing artifact-like augmentations to ensure a smooth transfer from synthetic to real domain.}  
    \textcolor{black}{(\textbf{3}) Through rigorous experiments, we demonstrate that our model performs robust gaze estimation even in the presence of noisy data. Given the ability of our model to learn the critical regions of the eye, our results set new state-of-the-arts on two benchmark gaze estimation datasets (Eyediap and UTMultiview) and achieves competitive performance on another dataset (MPIIGaze). We then validate our design choices through detailed ablation studies and exploring several variants of our proposed solution. To encourage reproducibility and contribute to the field, we make our code public at: \href{https://github.com/z-mahmud22/MSGazeNet}{https://github.com/z-mahmud22/MSGazeNet}.} 

The rest of this paper is organized as follows. In the next section, we describe the related work in the field. Following, we present our method, including the network and synthesized dataset. Next, we describe the experimental setup and implementation details. This is then followed by detailed experimental results and various sensitivity/ablation studies. Lastly, we summarize our work and discuss the potential future research directions.

\begin{table*}[t]
\footnotesize
\setlength
\tabcolsep{2pt}
\centering
\caption{An overview of existing gaze estimation methods.}

\begin{tabular}{l l l l l l}
 \hline 
 \textbf{Year} &
 \textbf{Method} &
 \textbf{Dataset} &
 \textbf{Input} & \textbf{Feature Extractor} & \textbf{Regressor} \\
 \hline\hline
 
 2015 & Zhang et al. \cite{lenet} & MPIIGaze, UTMultiview & Image, pose & LeNet & FC \\

 2017 & Zhang et al. \cite{mpii} & MPIIGaze, UTMultiview & Image, pose & VGG-16 & FC\\
 
 2018 & Park et al. \cite{gazemap} & Eyediap, MPIIGaze & Image & Hourglass+DenseNet & FC \\

 2018 & Park et al. \cite{ETRA} & Columbia \cite{columbia}, Eyediap, MPIIGaze, UTMultiview & Image & Hourglass & SVR \\

 2018 & Yu et al. \cite{clgm} & Eyediap, UTMultiview & Image, pose & 4 Layers CNN & FC \\
 
 2019 & Yu et al. \cite{imfew} & Columbia, MPIIGaze & Image & VGG-16 & FC \\
 
 2019 & Wang et al. \cite{bayes} & Columbia, Eyediap, MPIIGaze, UTMultiview & Image & Bayesian CNN & FC \\

 2020 & Yu et al. \cite{cvpr20} & Columbia, Eyediap, UTMultiview & Image & ResNet & FC \\
 
 2022 & Ghosh et al. \cite{mtgls} & Columbia, Gaze360 \cite{gaze360}, MPIIGaze & Image & ResNet-50 & FC \\
 
 2022 & Mahmud et al. \cite{mine} & Eyediap & Image & U-Net+Multistream VGG-16 & FC \\

 2023 & Cai et al. \cite{source} & Eyediap, ETH-XGaze \cite{ethxgaze}, Gaze360, GazeCapture \cite{gazecapture}, MPIIGaze & Image & Real-ESRGAN \cite{real-esrgan}+ResNet-18 & FC \\

 2023 & Jin et al. \cite{kappa} & Eyediap, MPIIGaze & Image, pose & VGG-16 & FC \\

 2023 & Jindal et al. \cite{gazeclr} & EVE \cite{eve}, Columbia, MPIIGaze & Image & ResNet-18 & FC \\

 \hline
\end{tabular}

\end{table*}

\section{Related Work}

Gaze is often represented as a 2D screen coordinate representing the point of gaze (PoG), or an angular vector representing the gaze direction by pitch and yaw angles.  
Vision-based methods primarily fall under three categories which are feature-based, model-based, and appearance-based. 
Feature-based approaches \cite{featurem1,huang,xiong} use the geometric shape of the eye to extract hand-crafted features such as eyeball centre, radius, pupil centre, eye corners from eye images which are then used in light-weight machine learning models to regress gaze. 
These methods were generally used prior to emergence of deep learning solutions. 
Model-based methods \cite{wood2014eyetab,wood20163d,wang2017real} aim to fit 3D deformable eye region models to eye images. Both classical machine learning and more recent deep learning techniques have been used in this category of literature.
Appearance-based methods \cite{unityeyes,mpii,clgm,gazemap} aim to learn a direct mapping of gaze from input eye images. These methods, which our paper also falls under, mainly rely on deep learning models. Following we present the related work under appearance-based gaze estimation methods. In particular, we review prior works on supervised learning, domain adaptation methods, as well as few-shot, semi-supervised, and self-supervised solutions. \textcolor{black}{Lastly, since segmentation of eye regions is used in our study, we also briefly review the works in this area.}

\subsection{Supervised Methods}
A multimodal network was proposed in \cite{lenet} where a LeNet type architecture was adopted to perform gaze estimation from eye images and head pose. 
In a subsequent work \cite{mpii}, a VGG type architecture was extended into a multimodal model that also used eye images and head pose as multimodal inputs.  
In \cite{clgm}, a deep multitask network was proposed where the network aimed to learn eye gaze and eye landmarks from eye images and their corresponding head pose information. The work explored the correlation of eye landmarks and gaze direction and argued for the eye landmarks to provide information cues for gaze estimation. 
Landmark detection in the form of Gaussian probability heatmaps of landmark coordinates from synthetic images was proposed in \cite{ETRA} using a stacked hourglass network \cite{hourglass}. The predicted landmarks were then fed into a support vector regressor (SVR) \cite{svr} to estimate gaze. The proposed solution improved iris localization, eyelid registration, and gaze estimation accuracies in both cross-dataset and person-specific settings. In \cite{gazemap}, a pictorial representation of gaze was proposed, which was hypothesized to be an intermediate eye image representation. 
The proposed pipeline consisted of a stacked hourglass network \cite{hourglass} which was trained to predict the intermediate \textit{gazemaps} from eye images, followed by a lightweight DenseNet architecture \cite{densenet} to regress gaze from the gazemaps. 
Another multimodal approach was proposed in \cite{flame}, which used both RGB eye images and their corresponding eye landmark heatmaps for gaze estimation. The two inputs were processed via separate CNN encoders to extract features which were then concatenated along with head pose information, and subsequently fed into dense layers that output 3D gaze direction.

\subsection{Domain Adaptation Methods}
A synthetic dataset, \textit{SynthesEyes}, was published in \cite{syntheseyes} to perform both eye shape registration and gaze estimation. The dataset offers a wide variation of synthesized eye images in terms of head pose, gaze, and illumination conditions. It was shown that when used for pre-training, the dataset results in significant performance improvement in a cross-dataset setting using the network proposed in \cite{lenet}.
Following the prior work, \textit{UnityEyes}, a synthesis framework was developed in \cite{unityeyes}, that can render eye region images in real-time. The system can be used to generate large scale synthetic eye image datasets and their corresponding landmarks, along with eye gaze annotations. 
A synthetic dataset consisting of millions of images which were generated by the simulator was used to train a simple kNN algorithm, outperforming their previous work \cite{syntheseyes} in cross-dataset experiments.  
To minimize the domain gap between synthetic and real images, a generative adversarial network (GAN) was proposed in \cite{apple} that used unlabeled synthetic and real eye images. The network consisted of a refiner network that refined the synthetic images to make them more realistic through adversarial learning. These refined images were then used to train a simple CNN to estimate gaze, which outperformed previous state-of-the-arts by a large margin in cross-dataset settings. A further improvement was reported in \cite{kaist} which relied on bidirectional mapping between synthetic and real eye images by leveraging a cyclic image-to-image translation framework. Highlighting the key challenges in cross-domain gaze estimation, a domain generalization technique was proposed in \cite{puregaze} where gaze-irrelevant features such as illuminations and appearance factors were eliminated via self-adversarial learning to extract purified gaze-relevant features from facial images. The conducted experiments resulted in new state-of-the-art performances in cross-dataset settings across multiple gaze estimation datasets without any fine-tuning. \textcolor{black}{In \cite{source}, a source-free domain adaptation method was introduced to adapt a gaze estimator to an unlabeled target domain without any source data. The neural architecture consisted of a face enhancer model that generates high-quality input images for the gaze estimator, leading to reduced variance and uncertainty of gaze predictions in the target domain.}

\subsection{Few-Shot and Unsupervised Methods}

For person-specific gaze estimation, a differential CNN was proposed in \cite{differential}, which output the gaze difference between two eyes of the same subject. During inference, the network used a set of calibration samples from the same subject and predicted the gaze difference between the input image and the calibration samples as the estimated gaze. With the intent of making more personalized gaze networks with lower gaze error, a person-specific gaze estimation network was proposed in \cite{faze} that worked with only a few ($\le$ 9) calibration samples from the test person. The proposed solution disentangled appearance features, gaze, and head pose information from facial images using a disentangling encoder-decoder (DT-ED) \cite{faze}. The network took an RGB face image as input and the decoder mapped it to three latent space vectors, which corresponded to eye region appearance, gaze, and head pose information respectively. The gaze latent vector was then fed to dense layers in order to make the gaze prediction. The scarcity of calibration samples in few-shot person-specific gaze estimation was addressed in \cite{imfew} by generating more training samples via synthesis of gaze-redirected eye images from an available set of calibration samples. 
The framework relied on synthetic images, generated by \cite{unityeyes}, to learn the gaze redirection task. To better adapt to the real domain, the network was further trained with real images which were first redirected given a redirection angle, and supervised via gaze redirection loss. Following, an inverse redirection angle was applied to the gaze redirected images to reconstruct the original images which were supervised via a cycle consistency loss. \textcolor{black}{In \cite{kappa}, a Kappa angle compensation method was proposed to neutralize the ocular counter-rolling response (OCR). The normalization process of eye images naturally induces the OCR which redistributes the Kappa angle's pitch and yaw component. This method with a few calibration samples ($\le$ 9) from the test subject, regresses the Kappa angle to refine the estimated gaze.}  

A semi-supervised approach was presented in \cite{bayes} where a Bayesian convolutional neural network (BCNN) relied on both labeled and unlabeled eye images to perform gaze estimation along with appearance classification and head pose estimation. The framework also included an adversarial component where the gaze labeled images were used as source domain and the unlabeled images are used as the target domain. The framework used the labeled images to supervise the gaze estimator, while the gaze estimator aimed to learn person-invariant features to oppose the adversarial module. Eye region segmentation was performed as an auxiliary task in \cite{mine} and the output eye segments and raw eye images were used as inputs to a multistream network for gaze estimation. The multistream network consisted of three encoders to extract features separately from the three inputs and the eye image encoder was pretrained with self-supervised contrastive learning and then fine-tuned during the downstream gaze estimation task. A sensitivity experiment verified the stability of the proposed network while using very limited amount of labeled data. In \cite{mtgls}, a multitask network was developed to perform three auxiliary gaze-relevant tasks with limited supervision. Using off-the-shelf networks, \textit{psuedo-gaze}, eye orientation and head pose were extracted from large scale facial image datasets \cite{msceleb, vggface2} which were later used to train a CNN backbone for the auxiliary tasks. To minimize noise in the generated labels, a noise distribution model was also incorporated in the framework. The network was then fine-tuned for downstream gaze estimation. \textcolor{black}{A contrastive learning method, \textit{GazeCLR}, was proposed in \cite{gazeclr} which pre-trains a CNN encoder with both single-view and multi-view gaze samples. These different gaze samples help the network learn invariance and equivariance among gaze representations, improving the cross-domain gaze estimation performance.}

In \cite{cvpr20}, an unsupervised approach was proposed to learn low dimensional gaze representations by utilizing a gaze redirection network. The proposed pipeline used an image pair of the same eye with different gaze directions as inputs to two separate networks for representations to be learned. The output latent vectors and their difference were then used in a gaze redirection network to reconstruct the latent representation of one of the images by redirecting the other image based on the gaze difference. The entire pipeline did not require ground truth gaze labels while training. The trained gaze representation network was then calibrated with randomly selected labeled samples (10-100) from the training data.

\begin{figure*}[t]
    \centering
    \includegraphics[width=0.9\linewidth]{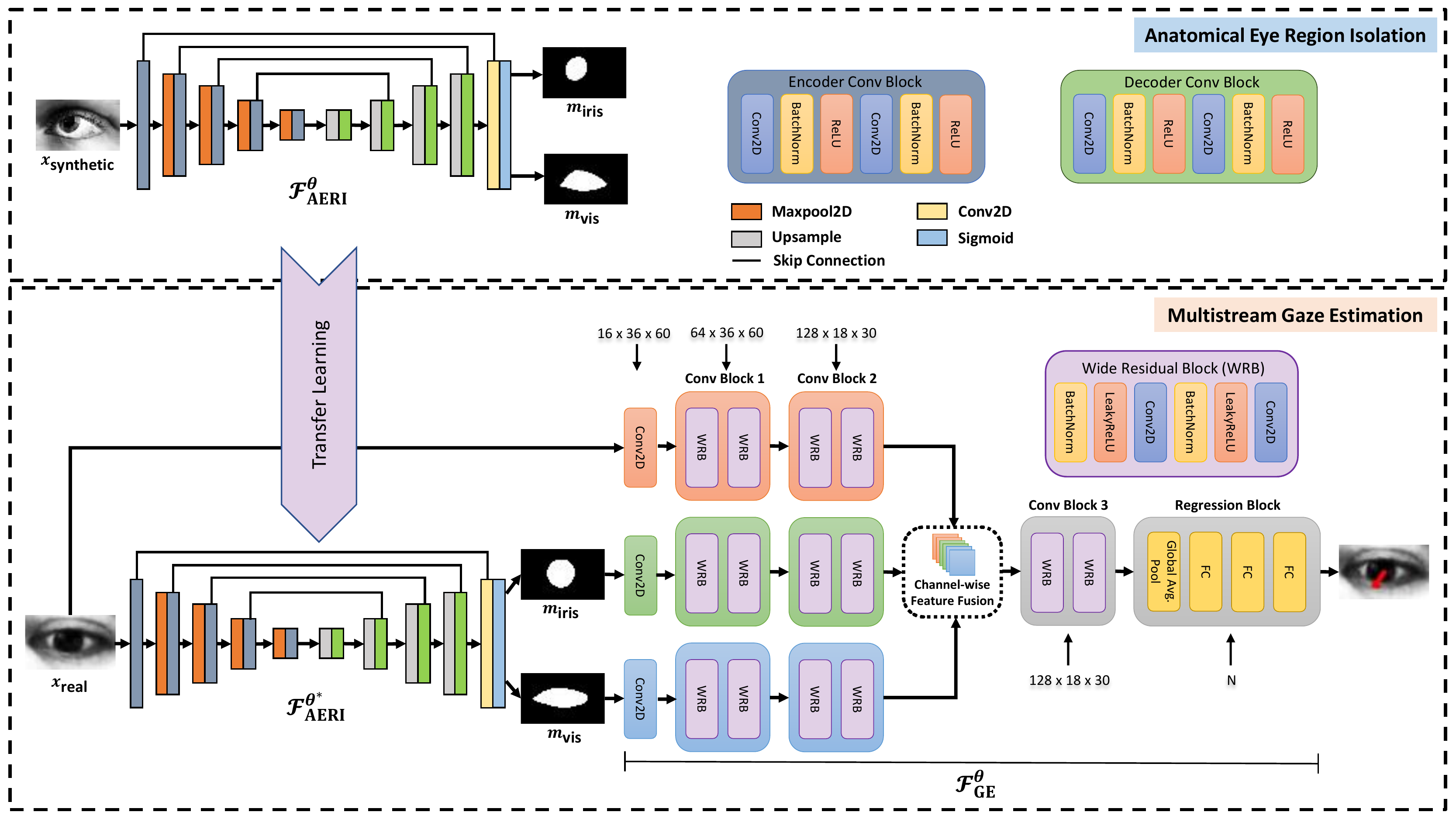} 
    \caption{Our proposed framework, MSGazeNet. First, we perform anatomical eye region isolation using a U-Net style network which we train using \textcolor{black}{the} synthetic dataset. Next, we perform gaze estimation using real-world eye images and their corresponding eye region masks as input to our multistream gaze estimation network.} 
    \label{fig:model}
\end{figure*}

\subsection{\textcolor{black}{Eye Region Segmentation}}
\textcolor{black}{Given that segmentation plays a role in our method, we briefly review the literature on this topic. 
In the context of segmentation focused on eye images \cite{casia, ubiris, ubirisv2}, a large-scale eye segmentation dataset, 
OpenEDS, was released in \cite{openeds}. Along with the dataset, some baseline experiments using deep convolutional encoder-decoder architectures \cite{segnet} were performed to set eye segmentation benchmarks.
Subsequently, an attention-based encoder-decoder network, Eyenet \cite{eyenet}, was proposed to perform multi-scale supervision during eye region segmentation. The neural architecture consisted of slightly modified residual units and two types of attention modules that applied attention on both channel and spatial dimensions. In \cite{kim2019eye}, a lightweight segmentation network based on MobileNetv2 \cite{mobilenetv2} was proposed which significantly reduced the processing time by utilizing depthwise convolution. A real-time segmentation network, RITnet, was released in \cite{ritnet} to segment eye images at 300 Hz. 
The proposed solution combined DenseNet with U-Net to create the architecture which was supervised via a weighted combination of three different loss functions. Semi-supervised approaches were explored in \cite{perryseg, semi-seg} where the solutions significantly improved the baseline performance with fewer annotated data and trainable parameters. In \cite{DAeyeseg}, three types of domain adaptation methods, supervised, semi-supervised, and unsupervised were explored using eye segmentation datasets collected from two different setups.}

\section{Method}
In this section, we first discuss the problem statement. This is followed by an overview of the proposed network, and detailed description of each component in our pipeline.

\subsection{Problem Statement}
Let's assume we have input $x$ which contains a grey-scale eye image. Our goal is to develop a model $\mathcal{F}$ which can reliably estimate the gaze parameters pitch ($\phi_{p}$) and yaw ($\phi_{y}$) angles. Pitch angles refer to the up/down eye movements and the yaw angles refer to the left/right eye movements. We hypothesize that in addition to learning the representation of the overall eye image, $x$, extracting information \textit{explicitly} from anatomical regions namely the visible eyeball $x_\text{vis}$ and the iris $x_\text{iris}$ would result in more informative features for estimating the final $\phi_{p}$ and $\phi_{y}$.

\subsection{Proposed Solution}

\noindent \textbf{Solution Overview.} To address the problem above, we design a network that first performs \textit{anatomical eye region isolation} in order to separate key geometric sections of the eye for explicit processing and representation learning. This critical step in our pipeline, as we will describe later, relies on transfer learning between simulation to real domains. Next, our pipeline uses all the available information, i.e., the original input along with the isolated eye regions, to perform representation learning followed by fusion, and eventually gaze estimation. An overview of our method is presented in Figure \ref{fig:intro}. In the following, we describe each of these components in our proposed method.

\vspace{3pt}

\noindent \textbf{Anatomical Eye Region Isolation.}
As touched upon above, a core component of our proposed method is the process of isolating different anatomical eye regions so that they could be individually used for gaze estimation. Here we first discuss our justification for including this step in our proposed method. 
Due to the inherent noise in real-world images, it is often quite difficult to recognize the gaze or orientation of the eye from raw images. Prior research \cite{clgm} suggests that gaze direction has a strong correlation with eye landmarks,  indicating that these landmarks could potentially contribute to gaze estimation as auxiliary information. However, obtaining such detailed and accurate landmarks is computationally expensive and susceptible to noise itself. Nevertheless, some methods use off-the-shelf landmark detectors which are primarily trained using synthetic data to extract eye landmarks. However, learning such high dimensional information is very challenging and those networks also suffer from the `synthetic to real' domain gap, which hinders the robustness of their landmark predictions. As a result, even though some methods \cite{ETRA, flame} use these landmarks in the form of heatmaps, there still remains considerable amount of noise in the training data.

In our proposed solution, as opposed to using the eye landmarks directly as auxiliary data, we propose and use the Anatomical Eye Region Isolation (AERI) network, to extract and isolates anatomical eye regions, namely the visible eyeball and the iris. We denote this network by $\mathcal{F}^{\theta}_\text{AERI}$ where $\theta$ are learnable parameters. $\mathcal{F}^{\theta}_\text{AERI}$ takes the eye image $x$ as input, and outputs $m_\text{vis}$ and $m_\text{iris}$, which are binary masks corresponding to $x_\text{vis}$ and $x_\text{iris}$ respectively. This design allows our pipeline to gain additional information about the orientation of the eye and key regions therein, without having to rely on the high-dimensional and noisy landmarks.

As shown in Figure~\ref{fig:model}, $\mathcal{F}^{\theta}_\text{AERI}$ uses a U-Net style architecture \cite{unet} with a two-channel output $[m_\text{iris},~m_\text{vis}]^T$. Thus, to train the network we require a dataset of $[x,~m_\text{iris},~m_\text{vis}]^T$ tuples. Since such a dataset does not exist and its collection from real images is quite difficult and time-consuming given the difficulty of recording eye images and isolating the anatomical regions for every image manually, we \textcolor{black}{introduce} a synthetic dataset. For this purpose, we rely on an eye image simulator UnityEyes \cite{unityeyes}, to \textcolor{black}{procure} the synthetic eye image dataset. The simulator can render synthetic eye images along with their detailed 2D landmark annotations and gaze labels in real-time. 
The simulator generates 32 landmarks for the iris region, denoted by $L_\text{iris} = [l_1,l_2,...,l_{32}]$, where $l_i \in \mathbb{R}^2$. Next, $m_\text{iris}$ is calculated  by:
\begin{equation}
    m_\text{iris} = {Bin_\text{enc}}\big(poly(L_\text{iris})\big),
\end{equation}
where $poly$ is a function that takes a series of landmark coordinates and creates a polygon by connecting them sequentially, and $Bin_\text{enc}$ is a binarizing operator which creates a binary mask by taking the enclosed area in its input and setting it to 1, while the outside is set to 0. 

\begin{figure}[t]
\centerline{\includegraphics[width=0.9\columnwidth]{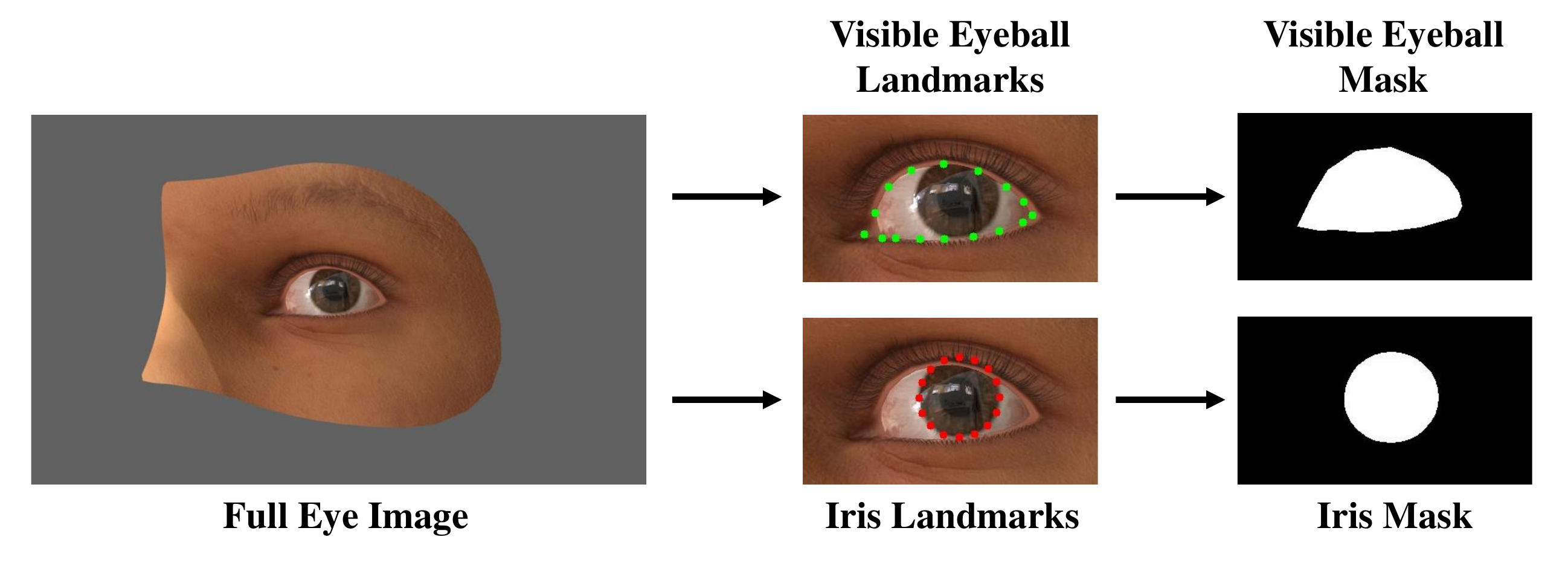}}
\caption{The mask generation pipeline. The visible eyeball landmarks are marked in green dots and the iris landmarks are marked in red dots. The binary masks are extracted from the corresponding landmarks.}
\label{fig:mask}
\end{figure}

The simulator also provides 16 landmarks for the interior region of the eye, i.e., the visible eyeball, plus 6 additional 2D coordinates for the caruncle region (inner corner of the eye). Here, we first average the 6 caruncle coordinates to create a single caruncle representative landmark, bringing the total landmarks for the visible eyeball to 17, as $L_\text{vis} = [l_1,l_2,...,l_{17}]$, where $l_i \in \mathbb{R}^2$. Next, similar to our approach for creating $m_\text{iris}$, we use: 
\begin{equation}
    m_\text{vis} = {Bin_\text{enc}}\big(poly(L_\text{vis})\big),
\end{equation}
to generate the visible eyeball mask. The process of obtaining $[x~m_\text{iris}~m_\text{vis}]^T$ is also depicted in Figure~\ref{fig:mask}.

Successive to \textcolor{black}{procuring} the synthetic dataset, we use it to train $\mathcal{F}^{\theta}_\text{AERI}$ using mean squared error (MSE) loss:
\begin{equation}
\label{eqn:segloss}
    \mathcal{L}_\text{AERI} = \frac{1}P \sum_{i=1}^P ||y_{i} - \hat{y}_{i}||^2 .
\end{equation}  
Here, $\mathcal{L}_\text{AERI}$ is the average MSE loss calculated between the predicted mask $\hat{y}_i = [\hat{m}_\text{iris},~\hat{m}_\text{vis}]^T$ and the ground truth mask $y_i = [m_\text{iris},~m_\text{vis}]^T$, for pixels $i \in P$.

As for the architectural details of the AERI network, as shown in Figure \ref{fig:model} (top), it consists of an encoder-decoder U-Net architecture. The architecture details presented in Table~\ref{table:unet_architecture} \textcolor{black}{are adapted from the classical U-Net model and are selected to maximize performance. }
The encoder contains 5 convolutional blocks (conv blocks) with 2$\times$2 maxpool layers in-between. The decoder consists of 4 conv blocks with 2$\times$2 upsampling layers in-between. A final 1$\times$1 convolution layer followed by sigmoid generates the network output. The feature maps from the upsampling layers of the decoder blocks are concatenated with their corresponding feature map outputs from the encoder blocks via skip-connections. Each conv block of both encoder and decoder consists of two 3$\times$3 convolution layers followed by batch normalization, zero-padding, and ReLU activation. The number of feature maps at the initial block of the encoder is 64, which is doubled following every maxpool to a maximum of 1024 feature maps at the last block of the encoder. Conversely, the number of feature maps in the initial decoder block is 1024, which is halved after every upsampling layer, reaching 64 at the final conv block.

\begin{table}[t]
\footnotesize
\centering
\caption{Architectural details for the anatomical eye region isolation network, $\mathcal{F}^{\theta}_\text{AERI}$.}
\label{table:unet_architecture}

\begin{tabular}{l|l|l}
 \hline
 \textbf{Modules} & \textbf{Parameters} & \textbf{Values} \\  
 \hline\hline
 Conv Block\textcolor{black}{(s)} & Input shape & 1$\times$36$\times$60 \\
 & \textcolor{black}{\# Encoder blocks}  & \textcolor{black}{5} \\
 & \textcolor{black}{\# Decoder blocks}  & \textcolor{black}{4} \\
 & \# Layers & 2 \\
 & Layer type & Conv2D \\
 & Kernel size & 3$\times$3 \\
 & Padding type & Zero-padding \\
 & Activation & ReLU \\
 \hline
 Downsample & \# Layers & 4 \\
 & Layer type & Maxpool2D \\
 & Kernel size & 2$\times$2 \\
 \hline
 Upsample & \# Layers & 4 \\
 & Layer type & Bilinear upsample \\
 & Scale factor & 2.0 \\
 \hline
 Output & \# Layers & 1 \\
 & Layer type & Conv2D \\
 & Kernel size & 1$\times$1 \\
 & Activation & Sigmoid \\
 \hline
 Full Network 
 & Batch size & 32 \\
 & Loss function & MSE \\
 & Optimizer & Adam \\
 & Learning rate & 0.00001 \\
 & Learning rate decay & 0.1 \\
 \hline

\end{tabular}
\end{table}

\noindent \textbf{Multistream Gaze Estimation.}
Successive to training the network described above, it is frozen and the weights are transferred to our final gaze estimation model. In order to ensure that domain-shift issues between the synthetic and real domains do not negatively impact the overall performance, we explore a variety of augmentations in order to allow the distribution of the synthetic dataset to become closer to that of the real datasets. This will be discussed in detail later in Section \ref{domainrand}.

We use the frozen AERI network, $\mathcal{F}^{\theta^ {*}}_\text{AERI}$ to generate $m_\text{iris}$ and $m_\text{vis}$. These two masks, along with $x$ are then used for gaze estimation. 
The layout of the gaze estimation network, $\mathcal{F}^{\theta}_\text{GE}$, is depicted in Figure~\ref{fig:model} (bottom) and the architecture is detailed in Table~\ref{table:gazenet_architecture}. The network consists of three parallel branches for the three different input streams. 
The first branch processes input eye image through a 
3$\times$3 convolution layer followed by two conv blocks with identical structures. 
We use wide residual blocks (WRBs) \cite{wideresnet} for the conv block architectures. Each conv block is a combination of two WRBs, denoted by \textit{B}(\textit{M}), where \textit{M} represents the list of kernel sizes of the convolution layers inside a WRB. The initial WRB is of type \textit{B}(3,1,3) which consists of two 3$\times$3 convolution layers followed by a 1$\times$1 convolution layer, while the subsequent WRB is of type \textit{B}(3,3). As mentioned earlier, the first conv block is followed by a second identical conv block. 
A similar architecture is used for the other two branches responsible for processing the eye anatomy mask streams. The feature maps obtained from the three branches are then fused in a channel-wise manner through concatenation, and subsequently fed to a single conv block (conv block 3) to extract combined representations.

\begin{table}[t]
\footnotesize
\centering
\caption{Architectural details for the proposed gaze estimation network, $\mathcal{F}^{\theta}_\text{GE}$.}
\label{table:gazenet_architecture}

\begin{tabular}{l|l|l}
 \hline
 \textbf{Modules} & \textbf{Parameters} & \textbf{Values} \\ 
 \hline\hline
 Conv2D & Input shape & 1$\times$36$\times$60 \\
 & Kernel size & 3$\times$3 \\
 & Padding type & Zero-padding \\
 & Output feature size & 16$\times$36$\times$60 \\
 \hline
 Conv Block(s)  & Architecture  & Wide Residual Network \\
 & \# WRBs & 2 \\
 & Block type & \textit{B}(3,1,3), B(3,3) \\
 & Activation & LeakyReLU \\
 & Dropout rate & 0.5 \\
 \hline
 Full Network & Widen factor & 4 \\
 & Depth & 16 \\
 & Batch Size & 32 \\
 & Loss function & MSE \\
 & Optimizer & Adam \\
 & Learning rate & 0.0001 \\
 & Learning rate decay & 0.5 \\
 \hline

\end{tabular}
\end{table}

\vspace{3pt}

\noindent \textbf{Gaze Regression.}
The combined feature embeddings from the multistream backbone network are passed through a global average pooling layer. The final feature embeddings are then flattened and fed to a set of 3 FC layers to regress the output gaze in the form of pitch ($\phi_{p}$) and yaw ($\phi_{y}$) angles. We use dropout with a probability of 0.25 and ReLU activation in between the FC layers, with the exception of the output layer. 

The gaze regression is supervised by the following loss:
\begin{equation}
\label{eqn:gazeloss}
\color{black}
    \mathcal{L}_\text{gaze} = \frac{1}{n} \sum_{i=1}^{n} ||g_{i} - \hat{g}_{i}||^2 ,
\end{equation} 
where $\mathcal{L}_\text{gaze}$ is the average MSE loss calculated between the predicted gaze \textcolor{black}{$\hat{g}_{i}$ = [$\hat{\phi}_{p}$, $\hat{\phi}_{y}$]} and the ground truth gaze \textcolor{black}{$g_{i}$ = [$\phi_{p}$, $\phi_{y}$]} for $i \in n$ sample images.

\subsection{Domain Randomization} \label{domainrand}
We apply a variety of augmentations during the training of $\mathcal{F}^{\theta}_\text{AERI}$ as well as $\mathcal{F}^{\theta}_\text{GE}$. Since the first component of the pipeline is solely trained on the synthetic domain, there remains a possibility for the network to not perform well in the real domain when integrated into the final MSGazeNet pipeline. To prevent this from happening, we adopt \textit{domain randomization} \cite{randomain} by applying visual transformations such as Gaussian noise, blur, and change of contrast to augment the simulated images so that they better reflect the distribution of the real-world images. The purpose of using this technique is to include a large number of relatively strong variations to the simulated images so that they represent the variations of the real-world datasets.
The parameters of all the applied transformations 
as detailed in Table~\ref{table:augmentation} 
are chosen via trial and error. During training, these transformations were randomly applied to augment the images. We apply Gaussian noise with a mean ($\mu$) of 0 and standard deviation ($\sigma$) in the range of [0 - 10.0]. We apply blurring operation with $\sigma$ in the range of [0 - 2.0] and a filter size of 3$\times$3. Downscale by a random factor in the range of [0-0.5] is applied followed by up-scaling the original image size, essentially quantizing the image with the original resolution. As another augmentation, we add random lines (0 to 2 lines). We also randomly change the contrast of the input images with a minimum intensity value, $r_{min}$ = [0-100] and a maximum intensity value, $r_{max}$ = [155-255]. Lastly, we randomly remove regions with a dynamic square kernel of size 1 to 10 pixels in both height (h) and width (w).
As we show later in the results section, we observe that applying strong augmentations enables both the anatomical eye region isolation network and the gaze estimation network to learn better representations for their respective tasks.

\begin{table}[t]
\small
\centering
\caption{Description of transformations.}
\label{table:augmentation}

\begin{tabular}{l c }
 \hline
 \textbf{Transformations} & \textbf{Parameters} \\  
 \hline\hline
 Gaussian noise & $\sigma$ = [0 - 10.0], $\mu$ = 0 \\ 
 Gaussian blur & $\sigma$ = [0 - 2.0], filter size = 3 $\times$ 3 \\
 Cutout & $\text{h,w}$ = [1 - 10.0] $\text{px}$ \\
 Downscale & [0-0.5]$\times$ \\
 Random lines & 0 to 2 \\
Contrast change & $r_{min}$ = [0-100], $r_{max}$ = [155-255] \\
 \hline

\end{tabular}

\end{table}

\begin{table*}[t]
\footnotesize
\centering
\caption{Overview of the \textcolor{black}{real-world datasets used in this paper}.}
\label{table:dataset}
    
    \begin{tabular}{c c c c c c c c}
     \hline
     \textbf{Name} & \textbf{\# Subjects} & \textbf{Size} & \textbf{Resolution} & \textbf{Head pose} & \textbf{Gaze target} &  \textbf{Illumination condition} & \textbf{Collection duration} \\  

     \hline\hline
     Eyediap \cite{eyediap} & 16 & 237 min & HD \& VGA & Continuous & Continuous & 2 & 2 days \\ 
     MPIIGaze \cite{mpii} & 15 & 213,659 & 1280$\times$720 & Continuous & Continuous & Daily life & $\sim$3 months \\
     UTMultiview \cite{utm} & 50 & 64,000 & 1280$\times$1024 & 8 + synthesized & 160 & 1 & 1 day \\
     \hline

\end{tabular}

\end{table*}

\section{Experiments}
In this section, we describe the datasets used in our work, followed by data normalization. Next, the evaluation protocol is described, followed by the description of the metric commonly used for quantifying gaze estimation performance. Lastly, the implementation details of our model are described.

\subsection{Datasets} 

\noindent \textbf{Real Datasets.}
We evaluate our model's performance on three public gaze datasets, \textbf{\textit{Eyediap}} \cite{eyediap}, \textbf{\textit{MPIIGaze}} \cite{mpii}, and \textbf{\textit{UTMultiview}} \cite{utm}. The overview of these datasets is shown in Table~\ref{table:dataset}. The Eyediap dataset contains 94 VGA and HD resolution videos captured from 16 subjects \textcolor{black}{among which 12 are male and 4 are female}. The dataset was collected in two different sessions for each subject. In one session, named the screen target session, small dots appeared on a screen which the users were asked to look at. In another session, called the 3D floating ball session, a ball was continuously moved in front of the user, which the user was asked to look at and follow. Both of these sessions included both static and mobile head pose scenarios. Although the dataset offers large variability in gaze and head pose, it contains similar illumination condition across all sessions. 

The MPIIGaze is a benchmark dataset for in-the-wild appearance-based gaze estimation. It has been collected from 15 subjects in an unconstrained manner over the course of several months. \textcolor{black}{Among these subjects, 9 are male, 6 are female, and 5 subjects wore glasses. The age of the participants ranges from 21 to 35 years old.} The dataset consists of 213,659 facial images of subjects during their everyday laptop usage and the corresponding ground truth gaze labels. The labels were collected using a custom software that triggered a sequence of 20 dots at every 10 minutes, which the users were asked to look at. The dataset contains significant variation in terms of appearance, head movement, gaze target, and illumination. 

The UTMultiview dataset consists of 160 discrete gaze targets, which contrasts the two datasets mentioned above given that they consist of continuous gaze targets. \textcolor{black}{There are 50 subjects (35 male and 15 female) who are aged between approximately 20 to 40 years old.} This dataset was collected in a laboratory environment and the collection procedure included 8 cameras to capture facial images of the participants from multiple viewpoints in a synchronized manner. The gaze labels were collected by instructing the participants to follow a visual target on a monitor.

\begin{figure*}[t]
\begin{center}
\subfigure{\includegraphics[width=4.0cm,height=4.0cm]{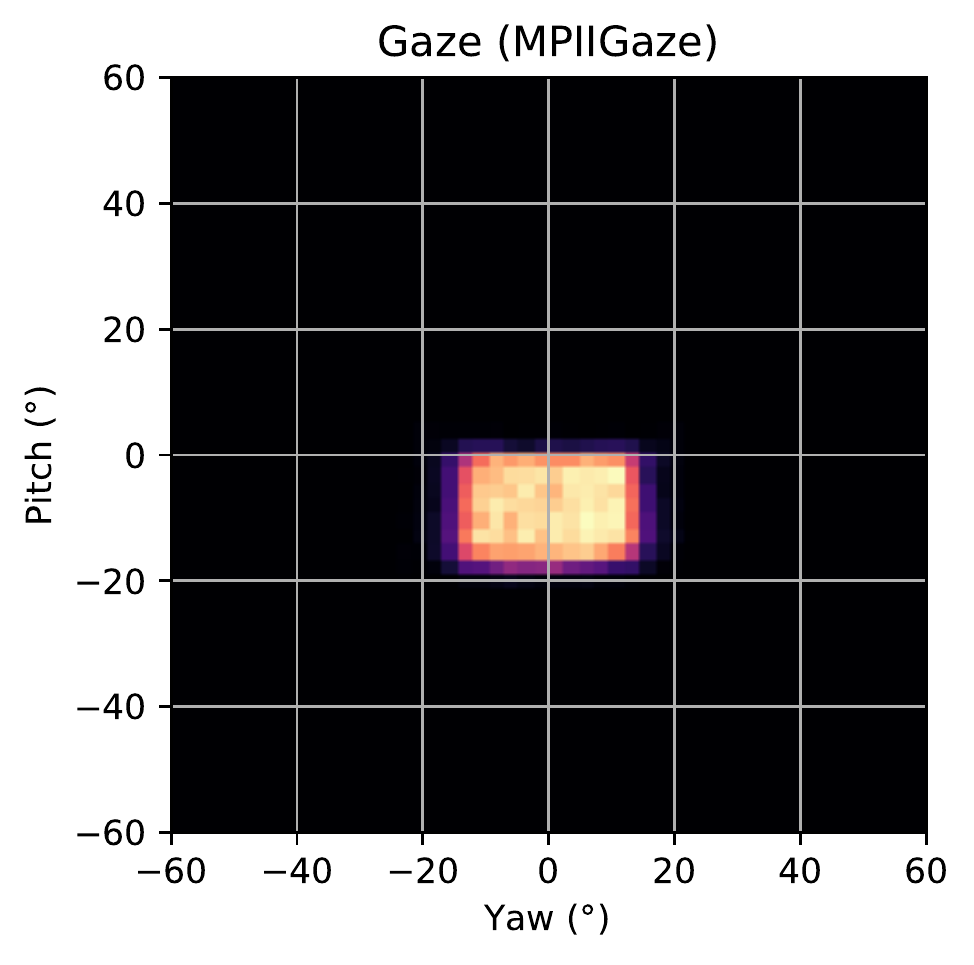}}
\subfigure{\includegraphics[width=4.0cm,height=4.0cm]{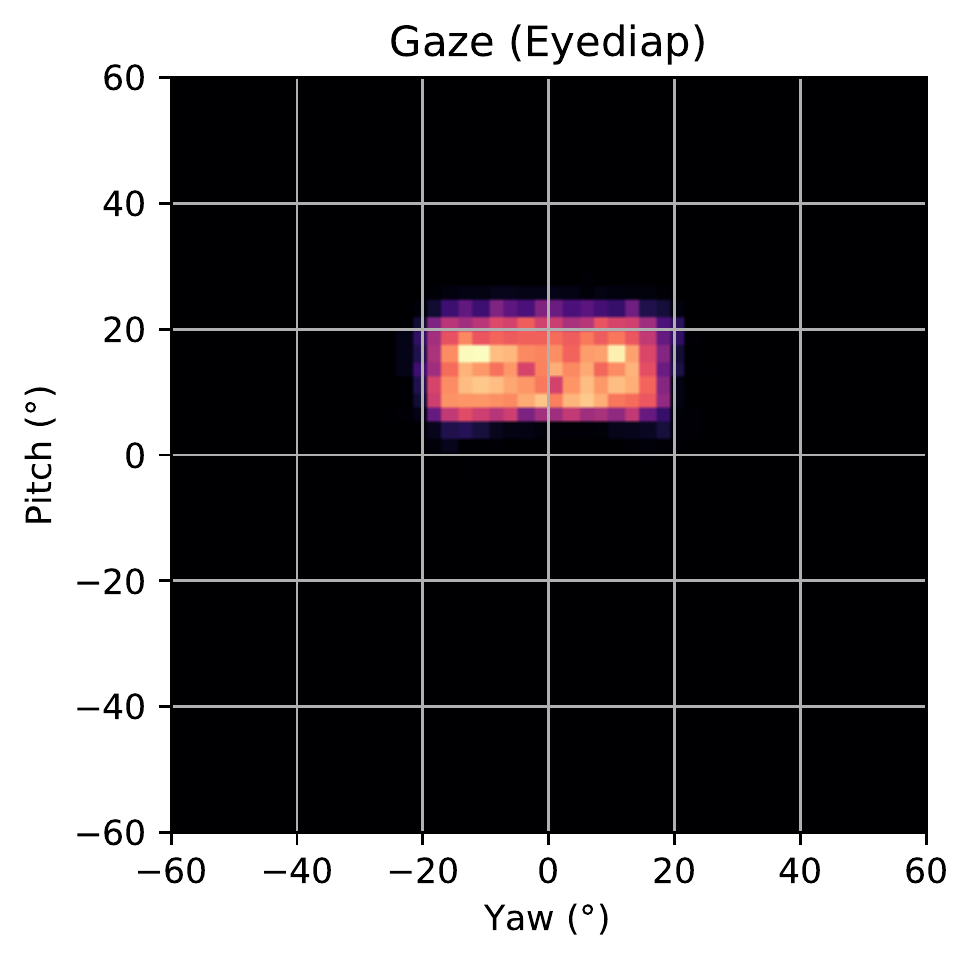}}
\subfigure{\includegraphics[width=4.0cm,height=4.0cm]{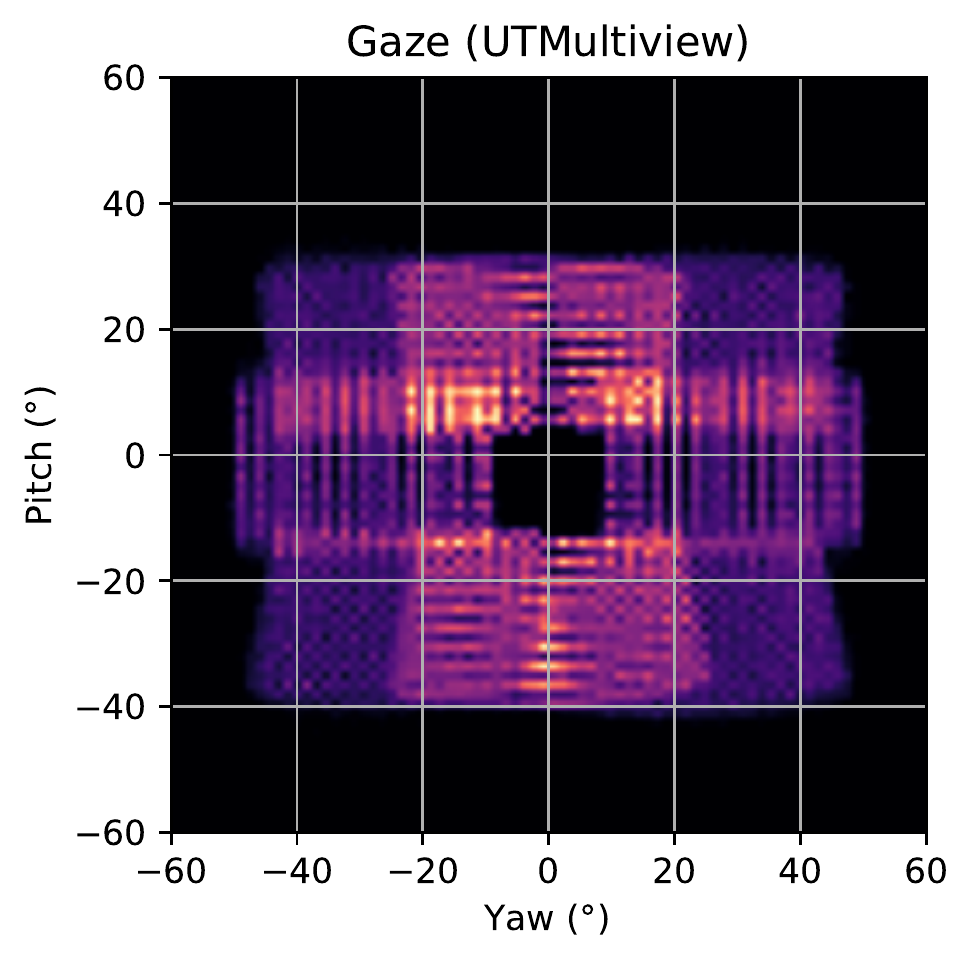}}
\subfigure{\includegraphics[width=4.0cm,height=4.0cm]{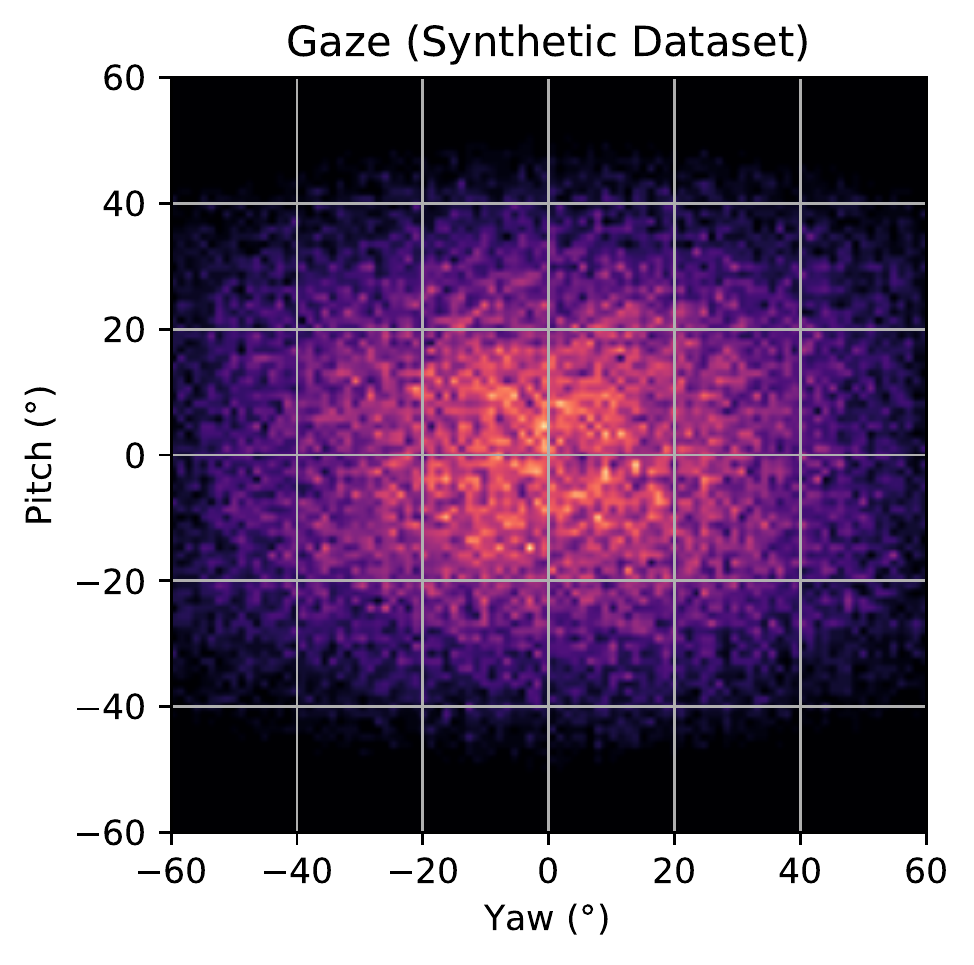}}

   \subfigure{\includegraphics[width=4.0cm,height=4.0cm]{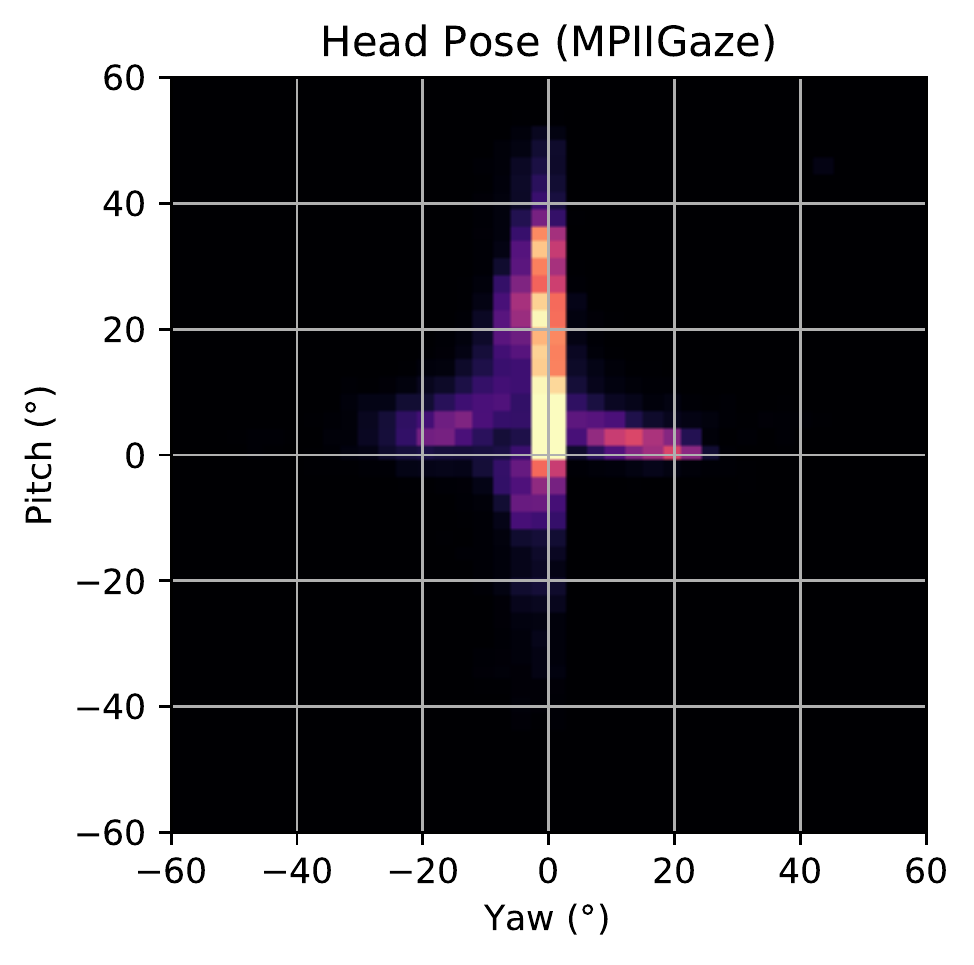}}
\subfigure{\includegraphics[width=4.0cm,height=4.0cm]{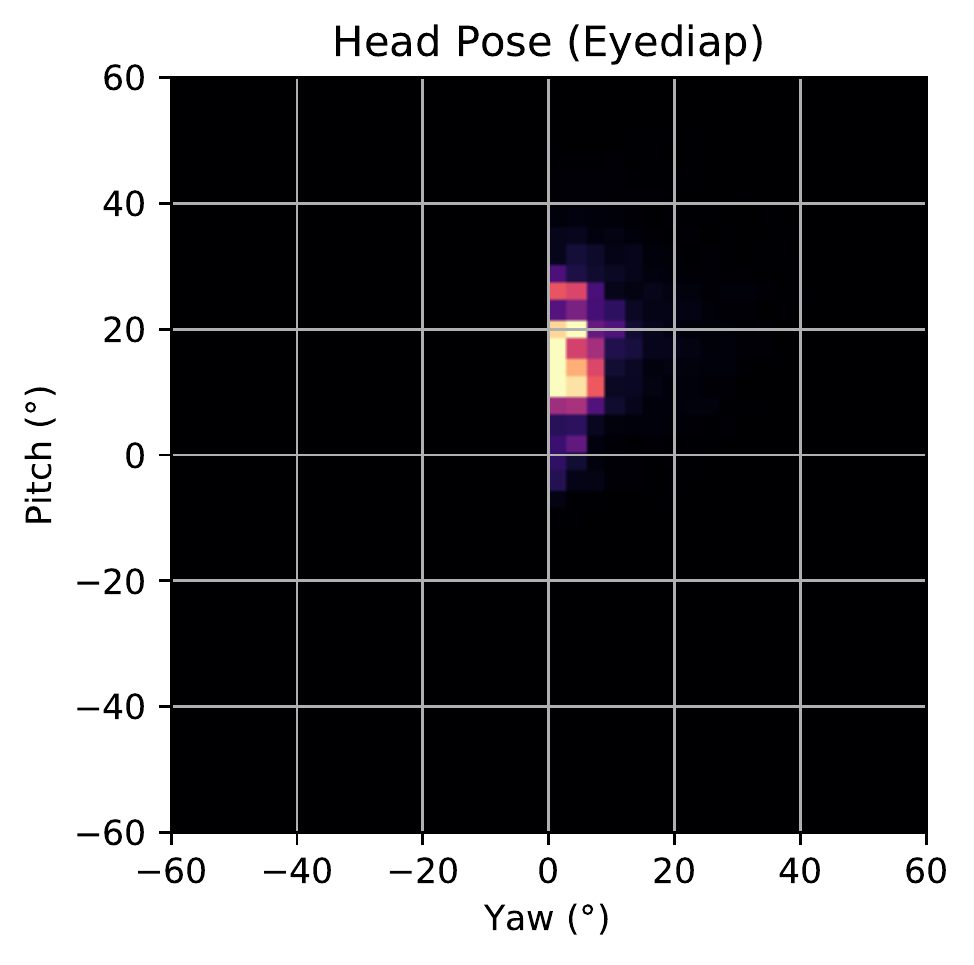}}
\subfigure{\includegraphics[width=4.0cm,height=4.0cm]{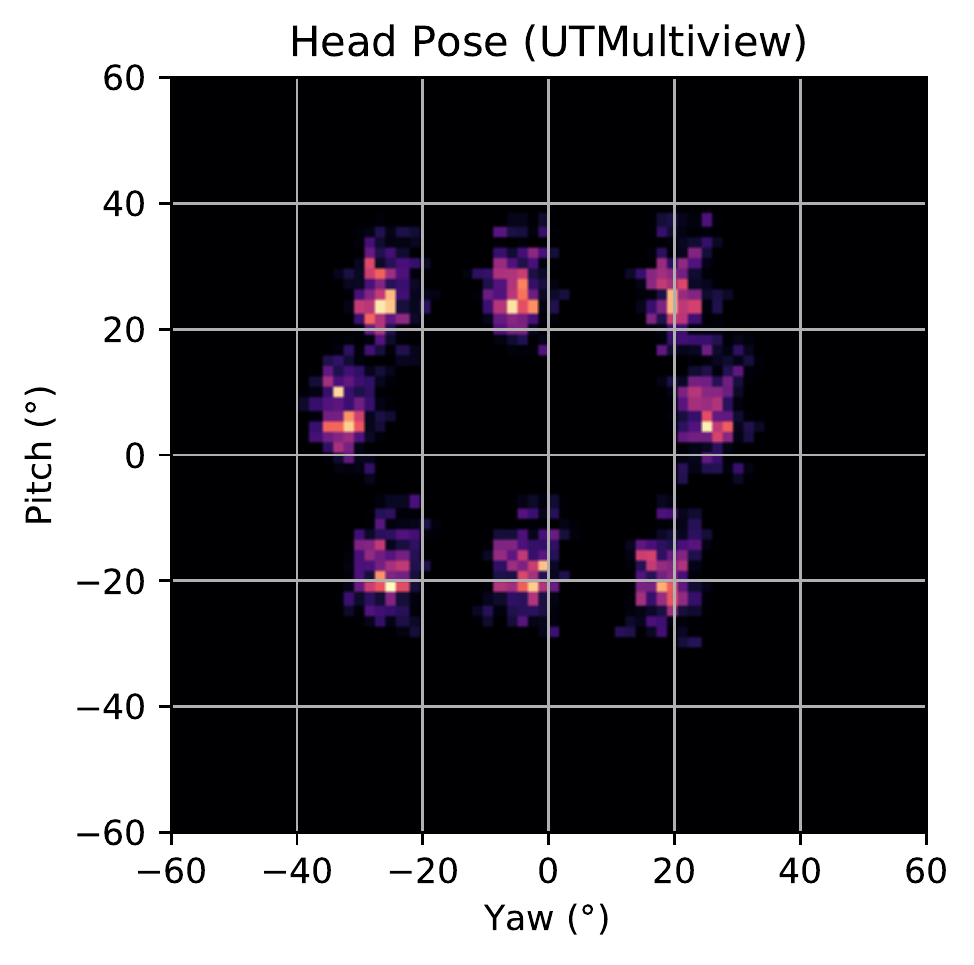}}
\subfigure{\includegraphics[width=4.0cm,height=4.0cm]{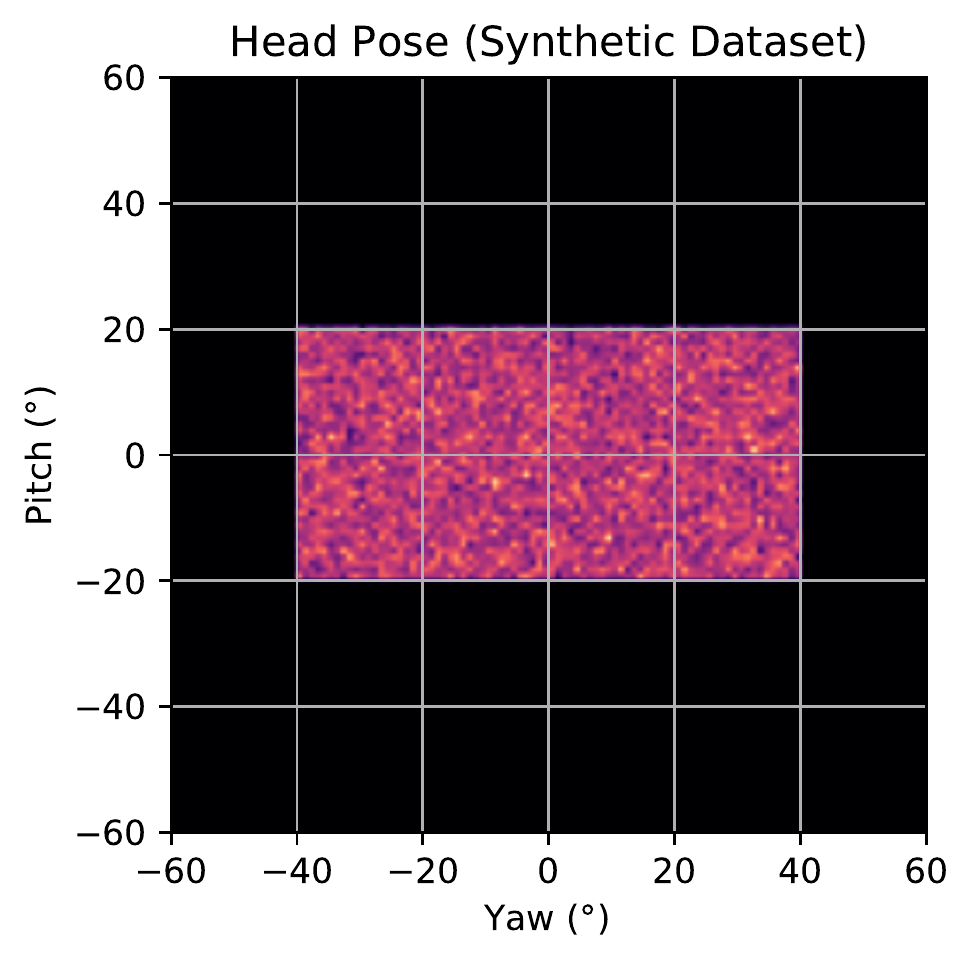}}

    \caption{\textcolor{black}{The heatmap distribution of the gaze angles (top row) and the head pose angles (bottom row) of all the datasets.}} 
\vspace{-0.75 cm}
\label{fig:heatmap_gaze_hp}
\end{center}
\end{figure*}

\noindent \textbf{\textcolor{black}{Synthetic Dataset.}}
In addition to the three datasets above, we use UnityEyes simulator \cite{unityeyes}, \textcolor{black}{to procure a synthetic dataset}, which consists of \textcolor{black}{80,000} synthetic eye images \textcolor{black}{with a resolution of 1280 $\times$ 768} and their corresponding masks to represent the iris and visible eyeball region of the eye. The images are \textcolor{black}{captured} by varying eye appearances, illuminations \textcolor{black}{(light intensity = [0.60 - 1.20])}, shapes, \textcolor{black}{headpose ($\phi_{p}$, $\phi_{y}$ = [$\pm$20.00$^{\circ}$, $\pm$40.00$^{\circ}$])}, and gaze \textcolor{black}{($\phi_{p}$, $\phi_{y}$ = [$\pm$49.49$^{\circ}$, $\pm$78.28$^{\circ}$])} directions, all of which are controllable parameters in the simulator. The dataset contains images from 20 virtual subjects out of which 5 are female and 15 are male. \textcolor{black}{For training, 60,000 of the synthetic images are used, while we set aside 20,000 for hold-out validation. \textcolor{black}{Finally, we present the distribution plots of both head pose and gaze for all the datasets used in this study in Figure~\ref{fig:heatmap_gaze_hp}}. We observe that the distributions of the synthetic dataset are wider than the three real-world datasets in terms of both gaze and head pose, making this dataset suitable for training purposes.}

\subsection{Data Normalization}
We normalize all the eye images and their corresponding gaze labels from Eyediap and UTMultiview datasets, as well as the images and masks from \textcolor{black}{the synthetic dataset}, according to \cite{revisit}. The MPIIGaze dataset directly provides normalized data. The normalization process requires 3D head pose and camera calibration matrix. A virtual camera is used to map the input images to a normalized space where a fixed distance of 600 mm is maintained between the eye centre and the virtual camera. The virtual camera is also rotated to point at the reference point (eye centre) as well as to prevent the freedom of head rotation in the roll axis to avoid any ambiguity.
Next, perspective transformation is applied to the images, followed by resizing to 60$\times$36 pixels. Finally, the eye images from all four datasets are converted to gray-scale, and then histogram equalized.

\subsection{Evaluation Protocol} \label{subsection:evaluation}

For gaze estimation, we use the VGA videos from the screen target session of the Eyediap dataset. The eye images are collected at every 15 frames from each recording to create a subset which we later use for training and testing. As there were no screen target session data for two subjects, the evaluation set contains data from 14 subjects rather than 16. The MPIIGaze dataset provides a subset of 45,000 eye images which is comprised of 1,500 left-eye and 1,500 right-eye images from each of the 15 subjects. 
The UTMultiview dataset contains both real-world and synthesized images for each participant. We use the real-world part which contains 64,000 eye images from 50 participants. We use the standard evaluation protocols for all the three benchmark datasets to be consistent with the previous methods \cite{mpii, gazemap, clgm, cvpr20}. We follow leave-one-subject-out evaluation for the MPIIGaze dataset, a 5-fold validation for the Eyediap dataset and a 3-fold validation for the UTMultiview dataset. Note that there is no subject overlap in the different folds for Eyediap and UTMultiview datasets.

\subsection{Evaluation Metric}
Following prior works \cite{mpii, gazemap, cvpr20}, the \textit{gaze angular error} is used to measure the gaze estimation performance. The gaze angular error $\delta$, is calculated between ground truth gaze \textcolor{black}{$g$} and predicted gaze \textcolor{black}{$\hat{g}$}. Before calculating $\delta$, the gaze angle values are converted to a three-dimensional vector $v_{g}$, of Cartesian coordinates by the following:
\begin{equation}
    v_{g} = T(g) = [-cos\phi_{p} sin\phi_{y}, \, -sin\phi_{p}, \, -cos\phi_{p} cos\phi_{y}] ,
\end{equation}
where $T(\cdot)$ represents the conversion between two coordinate systems. Hence, $\delta$ is calculated by:  \begin{equation}
     v_{g} = T(g), \quad v_{\hat{g}} = T(\hat{g}) ,
 \end{equation}
 and
 \begin{equation}
     \delta = arccos \ \frac{v_{g}^T \cdot v_{\hat{g}}}{||v_{g}|| \cdot ||v_{\hat{g}}||}.
 \end{equation}

\subsection{Implementation Details}
Both the anatomical eye region isolation and the gaze estimation networks are implemented using the PyTorch library. 
It should be noted that the setup of the regression block is optimized based on the datasets. While for all three datasets, three FC layers are used, the number of nodes vary. 
For MPIIGaze dataset, we use 512, 256, and 2 nodes for the three layers respectively. For Eyediap dataset, we use 256, 128 and 2 nodes. And lastly for UTMultiview dataset, we use 1024, 512 and 2 nodes. The AERI network is trained using an Adam optimizer \cite{adam} with an initial learning rate of $10^{-5}$ and a batch size of 32. The network is trained for 30 epochs via a step learning scheduler with a step size of 5 and a decay factor of 0.1. The gaze estimation network is also trained for 30 epochs using an Adam optimizer with an initial learning rate of $10^{-4}$ and a batch size of 32. At this step, a plateau learning rate scheduler is used with a decay factor of 0.5 and patience of 3. We train the entire pipeline (both the networks) using a single Nvidia 2080 Ti GPU. \textcolor{black}{The training time of the AERI network lasted for approximately 2 hours. The gaze estimator took about 0.5 hour per fold (out of 5) on the Eyediap dataset, 1.5 hours per fold (out of 15) on MPIIGaze dataset and 2.5 hours per fold (out of 3) on UTMultiview dataset. The inference time of our framework is 6 ms/image on GPU and 20 ms/image on CPU, making our model highly suitable for real-time systems.}

\section{Results}
In this section, we first present the quantitative results obtained by our model, and make comparisons against existing methods in the area notably the state-of-the-art. Following, we analyze the performance of our model in terms of robustness to noise. Next, we present qualitative results and visually demonstrate the performance of our method. This is followed by a series of thorough ablation studies to evaluate the impact of each major component of our solution. Finally we perform additional experiments on several network variants, feature fusion, and the impact of augmentations in our solution.

\subsection{Quantitative Results}
\textcolor{black}{At first, we present the quantitative results of our AERI network. To evaluate the performance of this module, we perform hold-out validation on the synthetic dataset. We use the MSE (Eq.~\ref{eqn:segloss}), and mean intersection over union (mIoU) as the evaluation metrics. It can be observed from Table~\ref{table:aeri_performance}, that we obtain low segmentation errors for both training and validation sets, while the overlap between the ground truth and predicted masks are quite high.}

\begin{table}[t]
\color{black}
\small
\centering
\caption{Performance of AERI.}
\label{table:aeri_performance}

\begin{tabular}{l c c}
 \hline
 \textbf{Dataset} & \textbf{MSE ($\downarrow$)} & \textbf{mIoU ($\uparrow$)}\\ 
 \hline\hline
 Train set & 0.011 & 0.875 \\ 
 Validation set & 0.098 & 0.892\\
 \hline

\end{tabular}

\end{table}

\begin{table}[t]
\footnotesize
\setlength
\tabcolsep{2pt}
\centering
\caption{Average gaze estimation error $\pm$ standard deviation for leave-one-subject-out evaluation on MPIIGaze dataset. The reported errors are in degrees ($^{\circ}$).}
\label{table:mpii_result}

\begin{tabular}{l l l l c}
 \hline
 \textbf{Method} & \textbf{Input} & \textbf{Feature Extractor} & \textbf{Reg.} & \textbf{Err.} \\
 \hline\hline
 Zhang et al. \cite{lenet} & Img+pose & LeNet & FC & 6.3 \\

 Zhang et al. \cite{mpii} & Img+pose & VGG-16 & FC & 5.5 \\
 
 Park et al. \cite{gazemap} & Img & Hourglass+DenseNet & FC & 4.56 \\
 
 Yu et al. \cite{imfew} & Img & VGG-16 & FC & 5.35 \\
 
 Wang et al. \cite{bayes} & Img & Bayesian CNN & FC & 4.3 \\
 
 Ghosh et al. \cite{mtgls} & Img & ResNet-50 & FC & \textbf{4.21}$\pm$1.90 \\
 
 MSGazeNet (Ours) & Img & U-Net+M.S. w. ResNet & FC & 4.64 $\pm$0.73 \\
 \hline
\end{tabular}

\end{table}

\begin{table}[t]
\footnotesize
\centering
\setlength
\tabcolsep{2pt}
\caption{Average gaze estimation error $\pm$ standard deviation for 5-fold evaluation on Eyediap dataset. The reported errors are in degrees ($^{\circ}$).}
\label{table:eyediap_result}

\begin{tabular}{l l l l c}
 \hline 
 \textbf{Method} & \textbf{Input} & \textbf{Feature Extractor} & \textbf{Reg.} & \textbf{Err.} \\
 \hline\hline
 Zhang et al. \cite{mpii} & Img+pose & VGG-16 & FC & 6.3$^{\ast}$ \\
 
 Park et al. \cite{ETRA} & Img & Hourglass & SVR & 11.9 \\
 
 Park et al. \cite{gazemap} & Img & Hourglass+DenseNet & FC & 10.3 \\
 
 Yu et al. \cite{clgm} & Img+pose & 4 Layers CNN & FC & 6.5 \\
 
 Wang et al. \cite{bayes} & Img & Bayesian CNN & FC & 9.9 \\
 
 Yu et al. \cite{cvpr20} & Img & ResNet & FC & 6.79 \\
 
 Mahmud et al. \cite{mine} & Img & U-Net+M.S. VGG-16 & FC & 6.34 \\
 
 MSGazeNet (Ours) & Img & U-Net+M.S. w. ResNet & FC & \textbf{5.86$\pm$0.80} \\
 \hline
\end{tabular}

$^{\ast}$ represents the error reported in \cite{clgm}
\end{table}

\begin{table}[t]
\footnotesize
\centering
\setlength
\tabcolsep{2pt}
\caption{Average gaze estimation error $\pm$ standard deviation for 3-fold evaluation on UTMultiview dataset. The reported errors are in degrees ($^{\circ}$).}
\label{table:utm_result}

\begin{tabular}{l l l l c}
 \hline 
 \textbf{Method} & \textbf{Input} & \textbf{Feature Extractor} & \textbf{Reg.} & \textbf{Err.} \\
 \hline\hline
 
 Sugano et al. \cite{utm} & Img+pose & Raster-scanner & kNN & 6.5 \\
 
 Zhang et al. \cite{lenet} & Img+pose & LeNet & FC & 5.9 \\
 
 Zhang et al. \cite{mpii} & Img+pose & VGG-16 & FC & 6.3$^{\ast}$ \\
 
 Yu et al. \cite{clgm} & Img+pose & 4 Layers CNN & FC & 5.7 \\
 
 Wang et al. \cite{bayes} & Img & Bayesian CNN & FC & 5.4 \\
 
 Yu et al. \cite{cvpr20} & Img & ResNet & FC & 5.52 \\
 
 MSGazeNet (Ours) & Img & U-Net+M.S. w. ResNet & FC & \textbf{5.30$\pm$0.57} \\
 \hline
\end{tabular}

$^{\ast}$ represents the error reported in \cite{clgm}
\end{table}

We follow the standard evaluation protocol \cite{mpii, gazemap, clgm, cvpr20} that was previously described in Section~\ref{subsection:evaluation} to quantify the performance of our method and compare our performance with the existing state-of-the-arts on different datasets. As our focus is on person-independent gaze estimation, we only consider and compare our results against published prior works which have also adopted this validation scheme. The results are reported in Table~\ref{table:mpii_result}, Table~\ref{table:eyediap_result}, Table~\ref{table:utm_result} in terms of angular error measured in degrees for MPIIGaze, Eyediap, and UTMultiview datasets, respectively. 
It should be noted that the result for \cite{mpii} on Eyediap and UTMultiview, has been taken from \cite{clgm} since the original paper \cite{mpii} did not report their performance for these datasets.
It can be observed that our proposed model, MSGazeNet, outperforms all the existing state-of-the-arts in both Eyediap and UTMultiview datasets while achieving competitive performance to \cite{mtgls} on the MPIIGaze dataset. On Eyediap dataset, we achieve a performance gain of 7.57$\%$ (6.34$^{\circ}$ $\rightarrow$ 5.86$^{\circ}$) over the existing state-of-the-art \cite{mine}. We also improve the current state-of-the-art \cite{bayes} on the UTMultiview dataset by 1.85$\%$ (5.4$^{\circ}$ $\rightarrow$ 5.3$^{\circ}$). Our results justify the significance of using anatomical eye region isolation along with raw eye images as a multistream input for gaze estimation. Despite not being very common practice in the literature, we also calculate and report the standard deviations for the performance of our method in each dataset. 
A considerably lower standard deviation compared to \cite{mtgls} on MPIIGaze (less than half) indicates the stability and certainty of our model.

\subsection{Robustness to Noise}
\begin{figure*}[t]
\begin{center}
\subfigure{\includegraphics[width=0.25\linewidth]{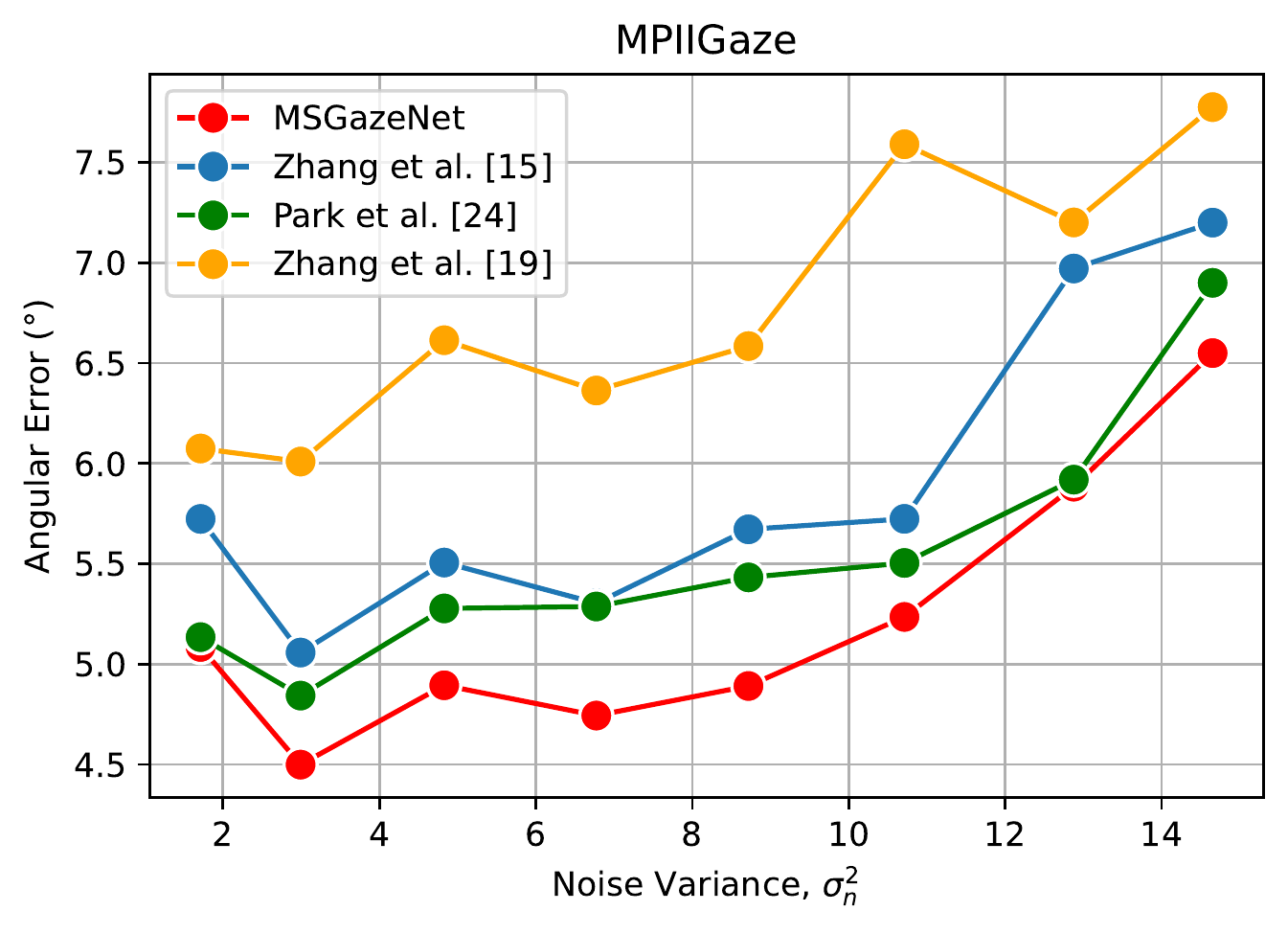}}
\subfigure{\includegraphics[width=0.25\linewidth]{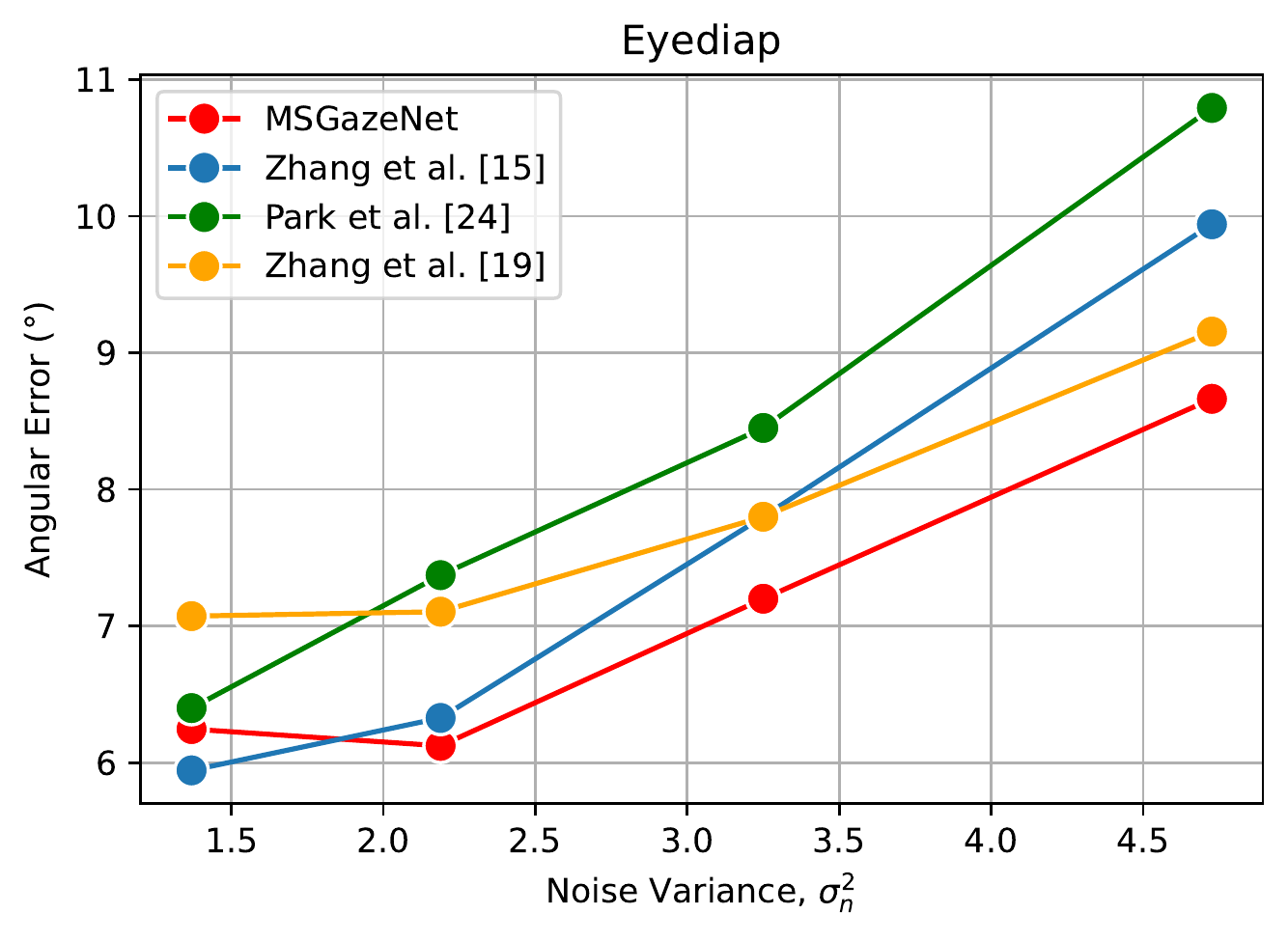}}
\subfigure{\includegraphics[width=0.25\linewidth]{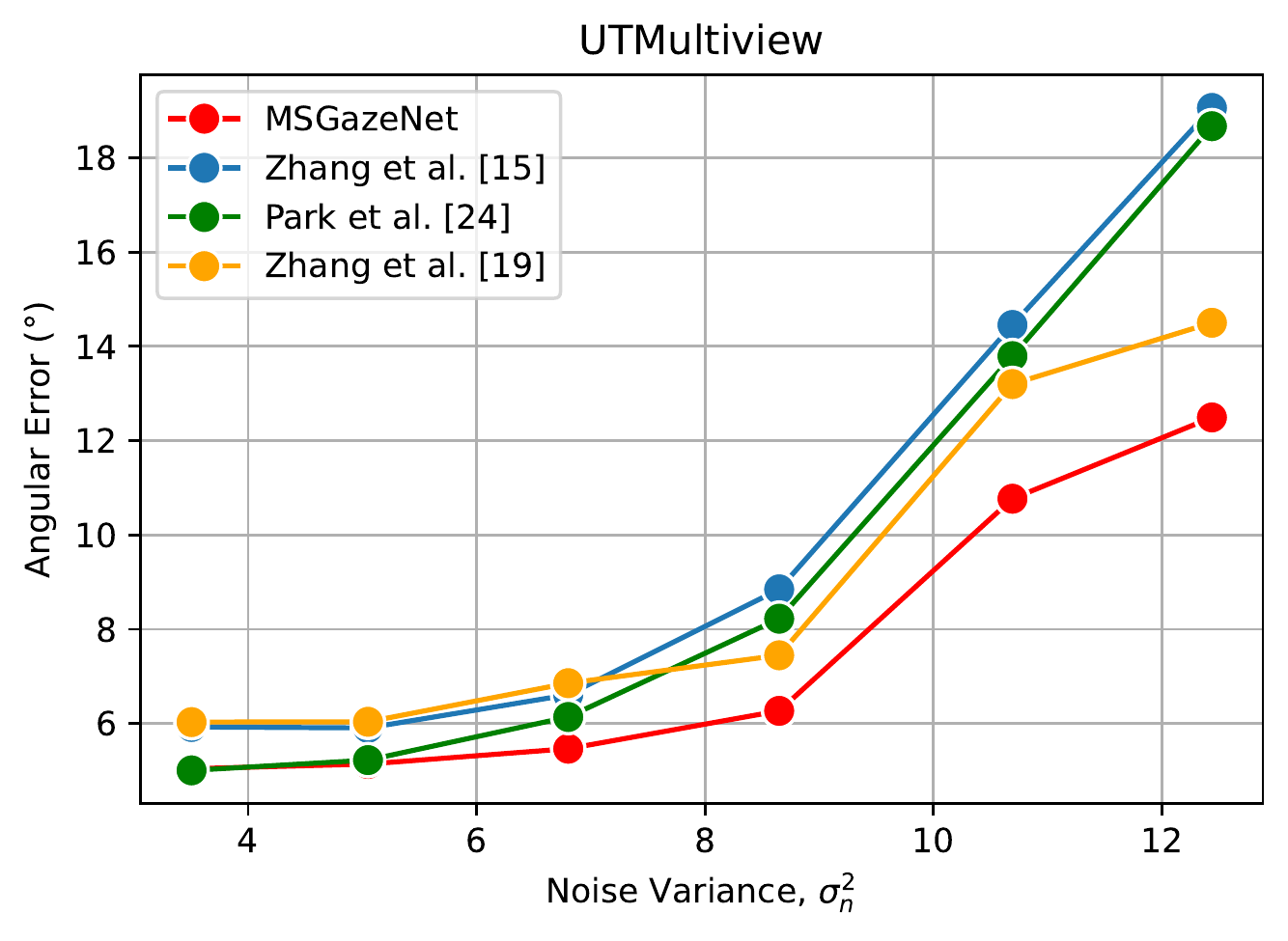}}
  
\caption{The outcome of the robustness analysis with respect to noise. Here we show the performance comparison of our proposed framework against \textcolor{black}{Zhang et al. \cite{lenet}}, Zhang et al. \cite{mpii} and Park et al. \cite{gazemap} in the presence of different amounts of noise.}
\label{fig:noiseplot}
\end{center}
\end{figure*}

In addition to comparing the results of our proposed method with existing state-of-the-art solutions, we also perform a robustness analysis to investigate the performance of our network in the presence of noisy data. In this analysis, following \cite{noise}, we first estimate the inherent noise in all the images from the previously mentioned real-world datasets (Eyediap, MPIIGaze, and UTMultiview) which we used for gaze estimation. The method estimates the variance of additive zero mean Gaussian noise in a given image, $I$ by convolving a noise operator, $N$ on the image. Following is the noise operator $N$, which is a mask of 3$\times$3 dimension as follows:
\begin{equation}
N =
\arraycolsep=1.0pt\def\arraystretch{1.0}
\begin{array}{|c|c|c|}
\hline
1&-2&1 \\
\hline
-2&4&-2 \\
\hline
1&-2&1 \\
\hline
\end{array}
\end{equation}
$N$ has zero mean and a variance of $36\sigma_{n}^2$. The noise variance, $\sigma_{n}^2$ of $I$ after applying $N$ can be computed by the following: 
\begin{equation}
   \sigma_{n}^2 = \frac{1}{36(W-2)(H-2)} \ \sum_{I} \ (I(x,y) \ast N)^2 ,
\end{equation}
where $W$ and $H$ are the width and height of the given image $I$ and $\ast$ represents the convolution operation at position $(x,y)$ of $I$. Next, we evaluate the performance of our network against the estimated noise variances in the input images and compare with prior works \cite{lenet}, \cite{mpii}, and \cite{gazemap}, for which public implementations and codes are available (at the time of performing this study, the codes for the other techniques were not made public). Our findings are plotted in Figure~\ref{fig:noiseplot} where we observe that as the amount of noise in the data increases, our model shows comparatively less deterioration than prior works, indicating better robustness to noise. \textcolor{black}{However, in the presence of significantly higher levels of noise, such as a noise variance exceeding 10 for the MPIIGaze and UTMultiview datasets and 3.5 for the Eyediap dataset, the performance of all the tested solutions, including ours, deteriorates.}

\begin{figure}[t]
\begin{center}
\subfigure{\includegraphics[width=0.33\columnwidth]{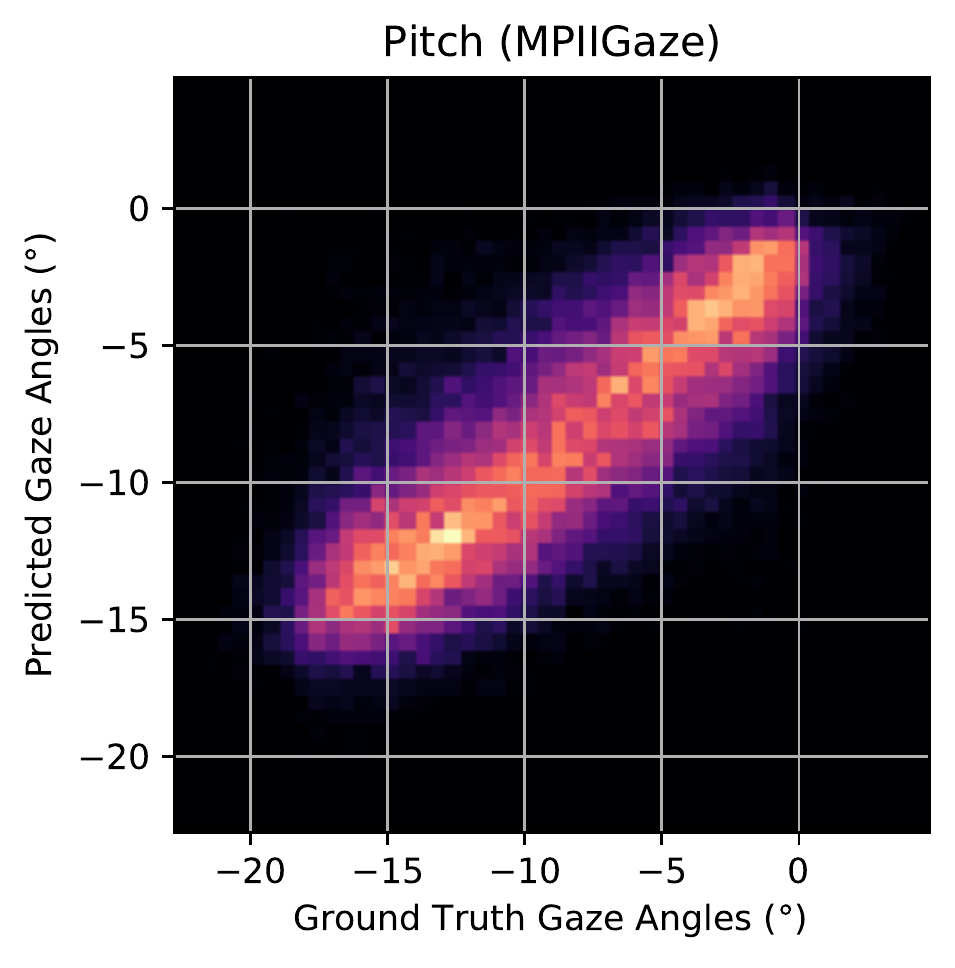}}%
\subfigure{\includegraphics[width=0.33\columnwidth]{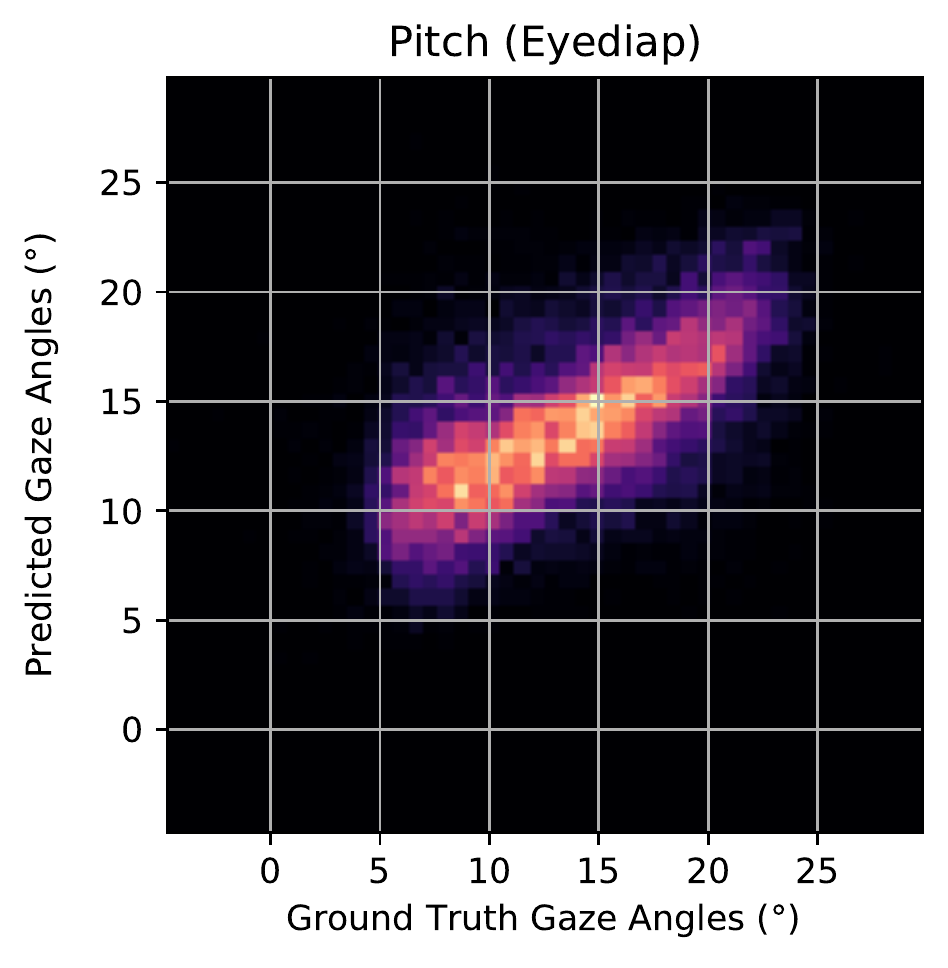}}%
\subfigure{\includegraphics[width=0.33\columnwidth]{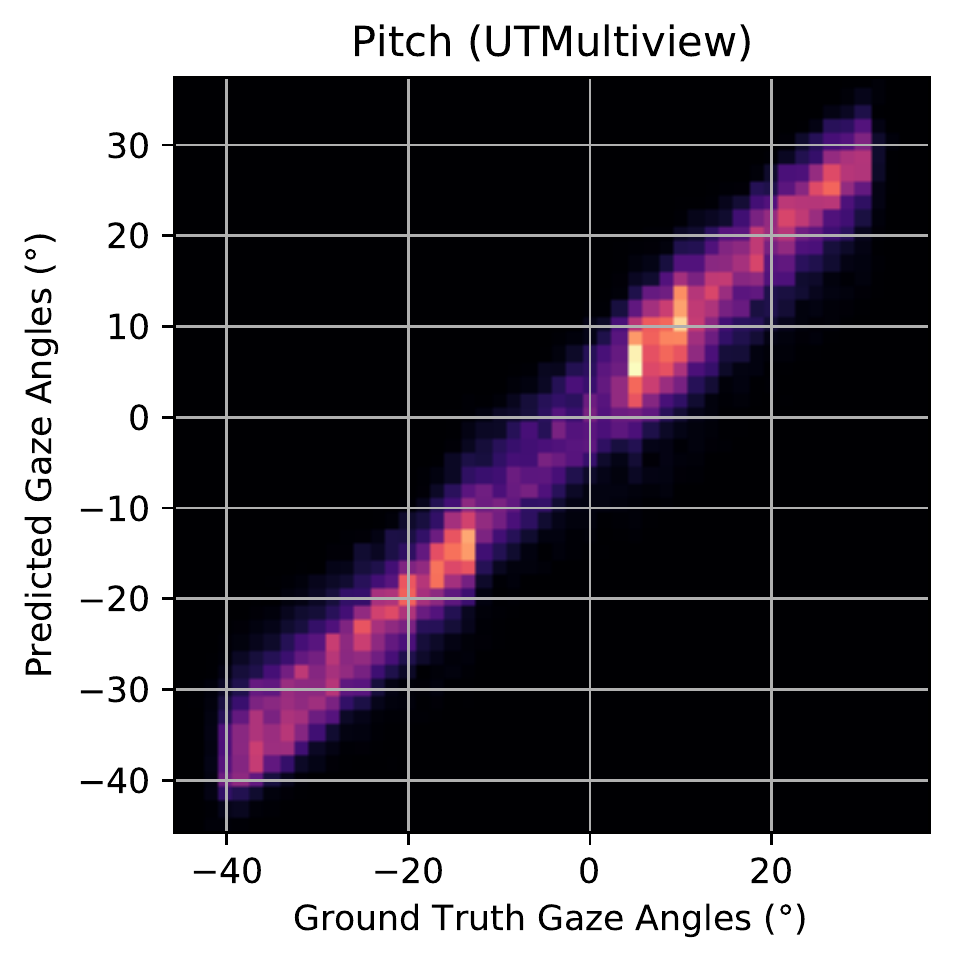}}%

   \subfigure{\includegraphics[width=0.33\columnwidth]{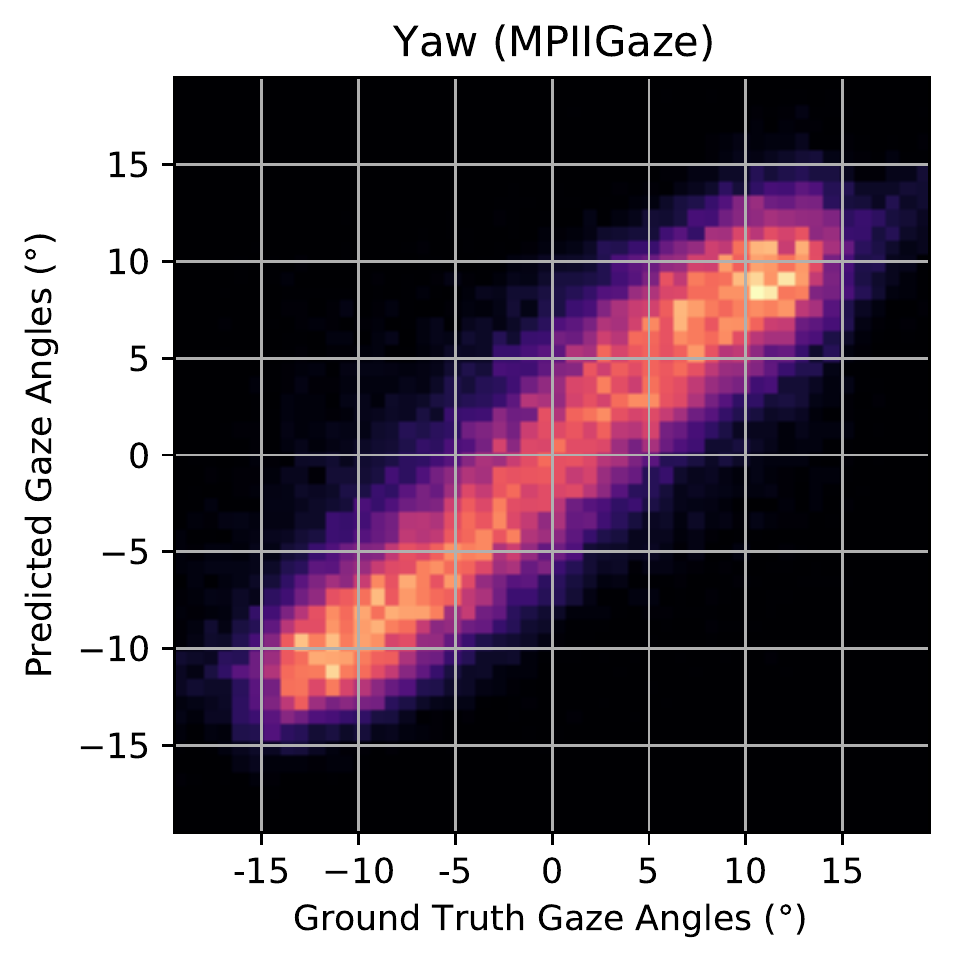}}%
\subfigure{\includegraphics[width=0.33\columnwidth]{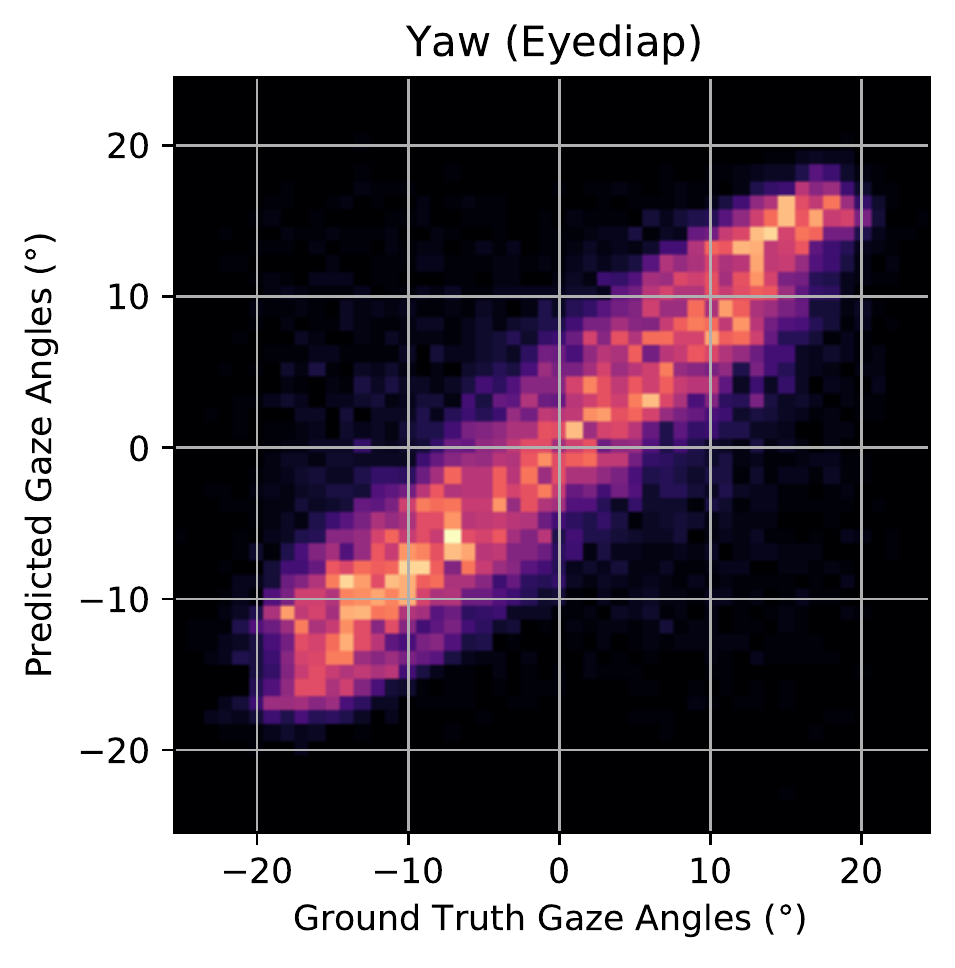}}%
\subfigure{\includegraphics[width=0.33\columnwidth]{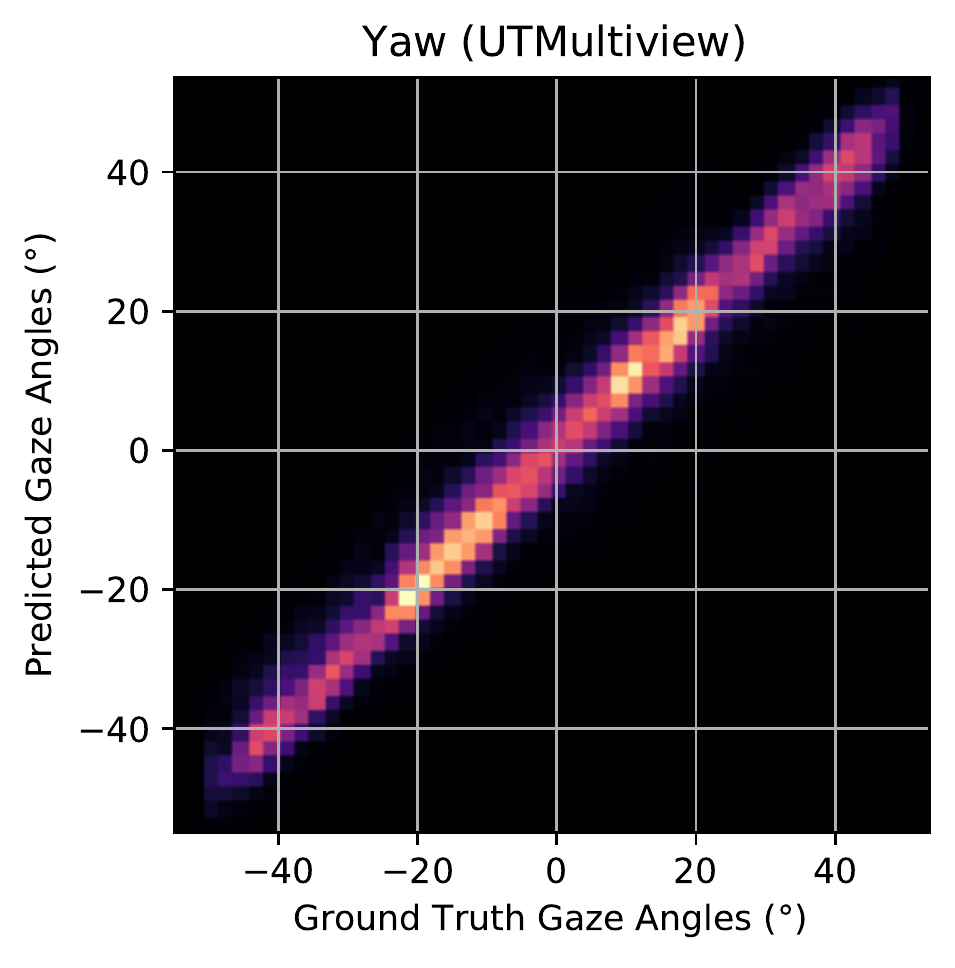}}%

    \caption{The heatmap distribution of the predicted gaze angles versus the ground truth gaze angles for pitch (top row) and yaw (bottom row).} 
\label{fig:heatmap}
\end{center}
\end{figure}

\begin{figure}[t]
    \centering
    \includegraphics[width=1\columnwidth]{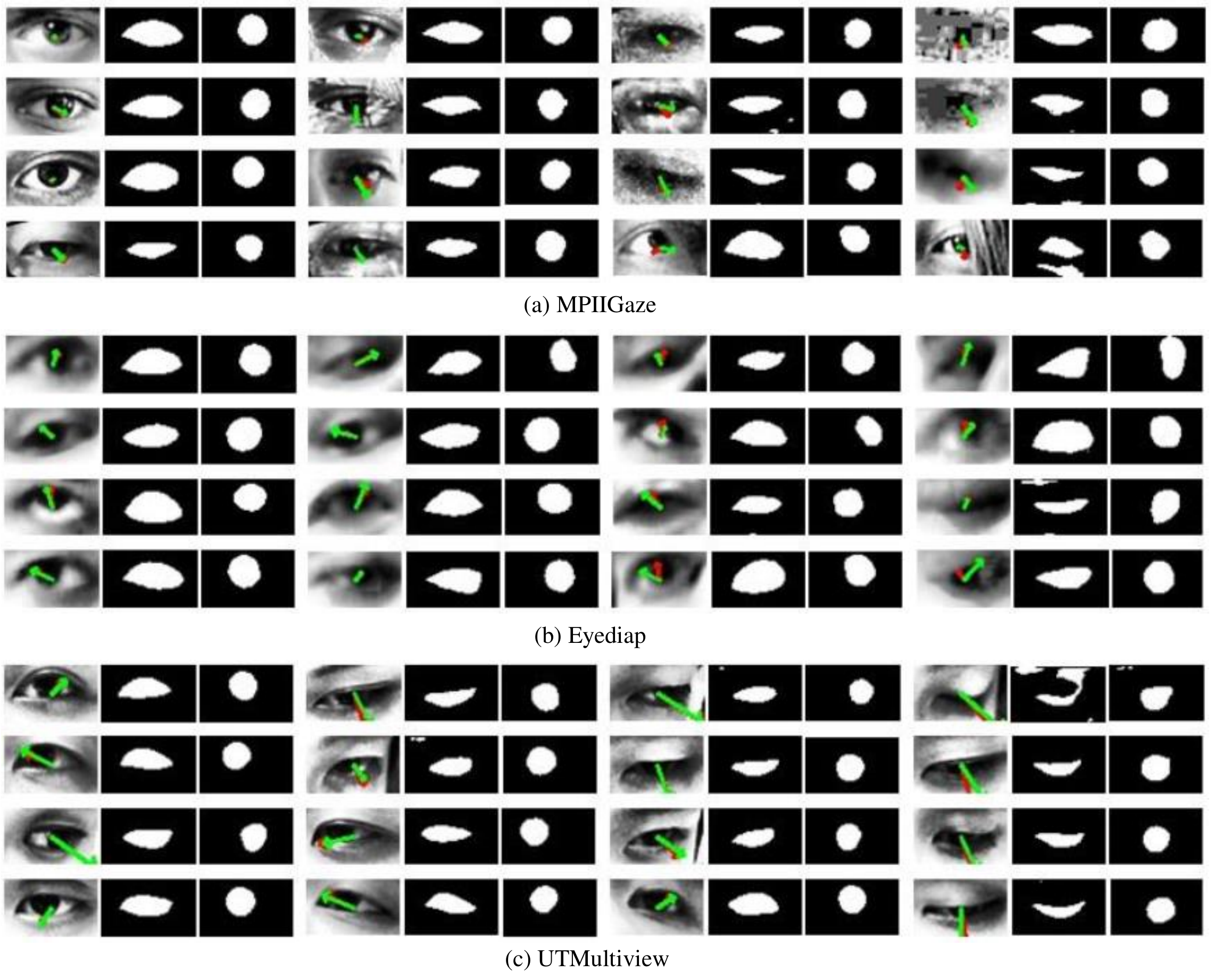} 
    \caption{The gaze predictions and eye region segmentation masks for samples from the three datasets. The predicted gaze angles are marked with green arrows and the ground truths are marked with red arrows.} 
    \label{fig:draw_gaze_segments}
\end{figure}

\subsection{Qualitative Results}
To carry out a qualitative analysis of the performance of MSGazeNet across different datasets, following \cite{flame}, we show a heatmap distribution of the predictions versus the ground truth gaze labels in Figure~\ref{fig:heatmap}. In the top row, the relationship between predictions and ground truths for the pitch angles ($\phi_{p}$) are illustrated while in the bottom row, the same relationship is shown for the yaw angles ($\phi_{y}$). The representations for all the three datasets exhibits almost a linear relationship between the predicted and the ground truth gaze angles. This further verifies the performance of our proposed framework. 

Next, we illustrate the output from our anatomical eye region isolation network and visualize the gaze angle predictions versus the ground truth gaze angles from the three benchmark datasets in Figure~\ref{fig:draw_gaze_segments}. 
The reported samples in each row are drawn from different participants within a dataset. 
The leftmost images of each row are comparatively less noisy with little to no occlusion while the rightmost images are more noisy or have more occlusions. 
From the figure, we can see that even in the presence of extreme noise, poor resolution, or partial occlusions the eye region masks can still capture the eye anatomy and eventually aid in estimating gaze. Expectedly, as the amount of noise and artifacts increase, the performance of our method does experience degradation, yet as shown in Figure~\ref{fig:noiseplot}, our model is relatively robust against such factors.

\subsection{Ablation Studies}
We conduct thorough ablation experiments to evaluate the design choices and various components in our proposed neural architecture. Following we describe these experiments and results.

\vspace{3pt}
\noindent \textbf{Impact of Each Stream.}
In our work, we hypothesize that using information from isolated anatomical eye regions in the model would aid gaze estimation. Here, we aim to explore the performance of our network using only one input stream at a time and subsequently adding the other streams for gaze estimation. To this end, we remove each branch of MSGazeNet, systematically, and evaluate the performance. First, we remove the two branches of the network responsible for learning representations from $m_{vis}$ and $m_{iris}$. 
The results presented in Table~\ref{table:exp_ablation} show that by removing the branches related to the anatomical eye regions, our model suffers performance degradation vs. the full model. Next, we remove the branch of the network that learns representations from the full eye image, 
and present the results in Table~\ref{table:exp_ablation}, where again we observe a performance drop. Subsequently, we remove only 
visible eyeball stream, 
followed by the removal of the iris stream 
to compare the gaze estimation performance using different combinations of input data. Similarly, both ablations result in performance drops, indicating that the full MSGazeNet is better capable of learning representations that are useful for gaze estimation.

\begin{table}[t]
\footnotesize
\centering
\caption{Ablation experiments on our proposed model. }
\label{table:exp_ablation}
\setlength
\tabcolsep{3pt}

\begin{tabular}{l| c c c}
 \hline
 \backslashbox{\textbf{Input}}{\textbf{Error ($^{\circ}$)}}{\textbf{Dataset}} & \textbf{MPIIGaze} & \textbf{Eyediap} & \textbf{UTMultiview} \\
 \hline\hline
 
 MSGazeNet (full network) & 4.64$\pm$0.73 & 5.86$\pm$0.80 & 5.30$\pm$0.57 \\

Eye regions removed & 4.95$\pm$0.69 & 6.19$\pm$0.78 & 5.51$\pm$0.72 \\ 

Eye image removed & 5.27$\pm$0.66 & 6.88$\pm$0.64 & 7.36$\pm$0.76 \\ 

Visible eyeball region removed & 4.83$\pm$0.65 & 6.27$\pm$0.78 & 6.97$\pm$0.79 \\ 

Iris region removed & 4.89$\pm$0.70 & 6.37$\pm$0.92 & 7.27$\pm$0.52 \\ \hline

Conv block 2 removed & 4.91$\pm$0.82 & 6.58$\pm$0.80 & 6.75$\pm$0.65 \\ 

Conv blocks 2 and 3 removed & 5.65$\pm$0.50 & 9.70$\pm$0.68 & 9.22$\pm$0.78 \\ \hline

\end{tabular}

\end{table}

\vspace{3pt}
\noindent \textbf{\textcolor{black}{Impact of AERI Network.}} 
\textcolor{black}{The isolation network for the anatomical eye region plays a vital role in our proposed framework. To evaluate the significance of this component, we conduct an experiment by substituting the AERI network with an equivalent counterpart. We utilize an off-the-shelf facial landmark extractor, OpenFace 2.0 \cite{openface2.0} to extract eye landmarks and subsequently create the eye region masks from the extracted landmarks. These masks and the corresponding eye images are used to perform gaze estimation, enabling us to conduct a comparative performance analysis. Due to the unavailability of face images, which is the requirement for facial landmark extraction via OpenFace 2.0, we could not use the UTMultiview dataset for this experiment. Our findings, presented in Table~\ref{table:exp_openface}, reveal that gaze estimation performance using the masks provided by the AERI network outperform the utilization of the OpenFace 2.0 toolkit. 
While both models are predicting outputs for real-world images, 
OpenFace 2.0 is estimating high dimensional landmark coordinates. 
Evidently, the predictions contain more noise in this case while we aim to simplify this process by predicting low dimensional binary masks for the eye regions and experimentally show that the predicted masks contribute better to accurately estimate gaze. This experiment illustrates that the AERI network can accurately capture crucial eye regions from real-world eye images to ensure more robust gaze estimation.} 

\begin{table}[t]
\color{black}
\footnotesize
\centering
\caption{\textcolor{black}{Performance comparison between the proposed network and the variant with OpenFace 2.0 on two datasets.}}
\label{table:exp_openface}
\setlength
\tabcolsep{3pt}

\begin{tabular} 
{l| c c c}
 \hline
 \backslashbox{\textbf{Method}}{\textbf{Error ($^{\circ}$)}}{\textbf{Dataset}} & \textbf{MPIIGaze} & \textbf{Eyediap} \\
 \hline\hline
 
 MSGazeNet (w/ AERI) & 4.64$\pm$0.73 & 5.86$\pm$0.80\\

 MSGazeNet (w/ OpenFace 2.0) & 4.98$\pm$0.71 & 6.31$\pm$0.72\\ 
 
 \hline
\end{tabular}

\end{table}

\vspace{3pt}
\noindent \textbf{Impact of Network Depth.}
Here, we investigate the performance of gaze estimation by varying the depth of our network. To this end, we systematically remove the convolutional blocks from all three branches and study the variation in performance.
First, we remove conv block 2 from each branch of our proposed network. Next, we further reduce the depth by removing conv block 3 from the remaining network. The results presented in Table~\ref{table:exp_ablation} reveal that the removal of the two conv blocks considerably deteriorates the performance.

\subsection{Network Variants}
Here, we create a number of variants of our model in order to further validate our network design choices. An element to our proposed neural network for gaze estimation is the usage of separate conv blocks to encode each individual input to extract optimal input-specific representations. Therefore we create three different variants of our network to explore the various possible encoding strategies for the three networks. These variants, which we describe below along with their performance, have all been trained and evaluated following the same setup and protocol as our proposed model.

First, we construct network Variant 1, 
in which a single series of conv blocks are used to encode all three inputs (eye image and the two anatomical eye regions). The results presented in Table~\ref{table:exp_networkvariants} show that for all three datasets, this strategy give results that are sub-par to our initial design, and in fact below state-of-the-art methods such as \cite{gazemap, bayes, mtgls} for MPIIGaze dataset, and \cite{mpii, clgm, bayes, cvpr20} for UTMultiview dataset. 

Next, we create Variant 2, in which the two anatomical eye region inputs are encoded together, while the full eye image is encoded separately \textcolor{black}{via a set of separate conv blocks}. 
Similar to Variant 1, we observe \textcolor{black}{the results from Table~\ref{table:exp_networkvariants} that indicates} deviating from our initial design choice of encoding each input separately degrades the performance, and places the results below \cite{gazemap, bayes, mtgls} for MPIIGaze dataset, and \cite{bayes, cvpr20} for UTMultiview dataset. 

Lastly, we explore the notion of extracting representations from the three branches with encoders that share weights. This strategy would considerably reduce the number of trainable parameters in the overall network. To this end, we create Variant 3 \textcolor{black}{that encodes each input stream separately with a series of conv blocks that share weights.} 
The results presented in Table~\ref{table:exp_networkvariants} show that this variant, similar to the other variants, degrades the performance, and falls below \cite{gazemap, bayes, mtgls} for MPIIGaze dataset, and \cite{bayes} for UTMultiview dataset.

\begin{table}[t]
\footnotesize
\centering
\caption{Performance comparison between the proposed network and the variants on all three datasets.}
\label{table:exp_networkvariants}
\setlength
\tabcolsep{3pt}

\begin{tabular}{l| c c c}
 \hline
 \backslashbox{\textbf{Method}}{\textbf{Error ($^{\circ}$)}}{\textbf{Dataset}} & \textbf{MPIIGaze} & \textbf{Eyediap} & \textbf{UTMultiview} \\
 \hline\hline
 
 MSGazeNet & 4.64$\pm$0.73 & 5.86$\pm$0.80 & 5.30$\pm$0.57 \\

 Var. 1 (1 encoding br.) & 4.85$\pm$0.76 & 5.97$\pm$0.85 & 5.94$\pm$0.89 \\ 

 Var. 2 (2 encoding br.) & 4.81$\pm$0.77 & 5.93$\pm$0.85 & 5.57$\pm$0.73 \\% [0.5ex]
 
 Var. 3 (3 encoding br., shared) & 4.83$\pm$0.66 & 5.97$\pm$0.91 & 5.45$\pm$0.57 \\

 \hline
\end{tabular}

\end{table}

\begin{table}[t]
\footnotesize
\centering
\caption{Performance comparison between the proposed network and variants with different feature fusion modules on all three datasets.}
\label{table:exp_fusion}

\begin{tabular}{l| c c c}
 \hline
 \backslashbox{\textbf{Method}}{\textbf{Error ($^{\circ}$)}}{\textbf{Dataset}} & \textbf{MPIIGaze} & \textbf{Eyediap} & \textbf{UTMultiview} \\
 \hline\hline
 MSGazeNet & 4.64$\pm$0.73 & 5.86$\pm$0.80 & 5.30$\pm$0.57 \\
 Late feature fusion & 4.87$\pm$0.82 & 6.12$\pm$0.79 & 6.08$\pm$1.02 \\
 Early feature fusion & 4.81$\pm$0.69 & 6.02$\pm$0.95 & 5.52$\pm$0.64 \\ 
 \hline
\end{tabular}

\end{table}

\subsection{Impact of Feature Fusion}
Given that our network uses separate conv blocks to extract features from each input, a feature fusion stage in our proposed gaze estimation architecture concatenates the individual feature maps in the channel dimension. Next, the fused features are fed to another conv block which is used to extract combined features. In this experiment, we create two baseline networks 
by modifying the position of the feature fusion stage to create late and early feature fusion variants. \textcolor{black}{To implement the late feature fusion, we place the feature fusion stage after conv block 3 instead of the original design. On the contrary for early feature fusion, we apply feature fusion prior to conv block 2.}
The results are presented in Table~\ref{table:exp_fusion}, where we observe that for all the datasets, we obtain the best performance from our proposed network compared to the network variants. Specifically, we see a performance drop in the range of 2.66\% to a maximum of $\sim$ 4\% in terms of mean angular error across different datasets using the different fusion strategies.

\begin{table}[t]
\footnotesize
\centering
\setlength
\tabcolsep{2pt}
\caption{Effect of domain randomization with different augmentation types on gaze estimation performance across all three datasets.}
\label{table:exp_aug}

\begin{tabular}{c c c c c c c c c}
 \hline
 \textbf{Noise} & \textbf{Blur} & \textbf{Cutout} & \textbf{Scale} & \textbf{Lines} & \textbf{Contrast} & \textbf{MPIIGaze} & \textbf{Eyediap} & \textbf{UTM} \\
 \hline\hline
 -- & -- & -- & -- & -- & -- & 4.85 & 6.14 & 5.53 \\
 \checkmark & -- & -- & -- & -- & -- & 4.73 & 5.98 & 5.42 \\
 -- & \checkmark & -- & -- & -- & -- & 4.77 & 6.14 & 5.39 \\
-- & -- & \checkmark & -- & -- & -- & 4.74 & 6.11 & 5.40 \\
-- & -- & -- & \checkmark & -- & -- & 4.76 & 6.02 & 5.42 \\
-- & -- & -- & -- & \checkmark & -- & 4.74 & 5.94 & 5.38 \\
-- & -- & -- & -- & -- & \checkmark & 4.73 & 5.97 & 5.45 \\
 \checkmark & \checkmark & \checkmark & \checkmark & \checkmark & \checkmark & 4.64 & 5.86 & 5.30 \\
\hline\
\end{tabular}
\end{table}

\subsection{Impact of Domain Randomization}
Domain randomization is an essential step in our solution for transferring the anatomical eye region isolation network, which was trained with synthetic and simulated data, into the real domain.
\textcolor{black}{This synthetic data does} not contain the various types of noise and artifacts which would normally be encountered with real datasets, in this case MPIIGaze, Eyediap, and UTMultiview.
Hence, to ensure a significant domain mismatch does not occur, we apply different strong augmentations when training the eye region isolation network. Here, we study the influence of the applied augmentations. To this end, we first train the network without any augmentations, and then add one augmentation at a time. In the end, all the augmentations are applied. \textcolor{black}{Throughout these individual steps, the size of the training data remained the same as the original 60,000 images.} The trained network is then integrated into the MSGazeNet framework. It can be observed from the results shown in Table~\ref{table:exp_aug} that when no augmentations are used, the gaze performance drops for all the datasets. This performance drop, should it have not been prevented, would have resulted in lower performances in comparison to \cite{gazemap, bayes, mtgls} for MPIIGaze, and \cite{bayes, cvpr20} for UTMultiview dataset.
Our experiments further show that adding `random lines' seems to be the most effective augmentation in the context of our study.
Nonetheless, the combination of all the augmentations yields the best results, pushing the performance of our model past the state-of-the-art on Eyediap and UTMultiview.

\section{Conclusion and Future Work}
In this work, we present MSGazeNet, a novel gaze estimator. Our solution performs person-independent gaze estimation, and consists of two integral parts namely the anatomical eye region isolation and multistream gaze estimation. The anatomical eye region isolation is a crucial component of our framework which is solely trained with synthetic data due to the scarcity of detailed and accurate eye region annotations in real-world gaze estimation datasets. To this end, we \textcolor{black}{procure} a synthetic dataset using UnityEyes eye-gaze simulator. Our dataset consists of \textcolor{black}{80,000} images along with the eye visible eyeball region and iris masks. This dataset is used to train a U-Net style model to isolate eye regions given an input eye image. To allow for this network to then be used for downstream integration into a model for real-world (not synthetic) gaze estimation, we perform domain randomization using a variety of artifact-like augmentations, which helps to narrow the domain gap. The eye region isolation network is then transferred into our gaze estimation pipeline which consists of a multistream architecture. The network takes raw real-world eye images along with the eye region masks to predict the gaze direction. We perform various experiments and demonstrate that our solution achieves strong results, achieving the state-of-the-art on Eyediap and UTMultiview datasets, and exhibiting competitive performance on MPIIGaze dataset. Our robustness experiments show that our model, is more robust in dealing with noise, in comparison to other methods. Detailed and thorough ablation studies and comparisons with several variants quantify the impact of each component in our network and validate our design choices.

In this study, we demonstrated the relevance and importance of using information pertaining to anatomical eye regions towards gaze estimation. This work can serve to motivate further research into using various regions of the eye for gaze estimation. We believe such approaches can lead to learning better and more generalized gaze representations. In addition to the above, a key area to investigate could be the implementation of semi-supervised learning to take advantage of large amounts of unlabeled real-world eye images while training the eye region isolation network. This could improve the eye region isolation module and further enhance the overall performance. Another important scope of research could be the consideration of using 3D representations of key eye regions (i.e. depth maps) which can be constructed from 3D landmarks. It is likely that the 3D representations would contain richer anatomical information about the eye in terms of eyeball curvature or iris contour, which would better aid gaze estimation. 

\noindent \textbf{Acknowledgment.} The authors would like to thank Innovation for Defence Excellence and Security (IDEaS) program for funding this project. The authors would also like to thank Dr. Dirk Rodenburg for his help throughout the project.

\bibliographystyle{IEEEtran}
\bibliography{references}

\end{document}